\documentclass{article}

\usepackage{iclr2026_conference,times}

\usepackage{latexsym}
\usepackage[T1]{fontenc}
\usepackage[utf8]{inputenc}
\usepackage{microtype}
\usepackage{inconsolata}
\usepackage{xcolor}
\usepackage{graphicx}

\usepackage{amsmath}
\usepackage{amsfonts}
\usepackage{booktabs}
\usepackage{array}
\usepackage{nicefrac}
\usepackage{adjustbox}
\usepackage{enumitem}
\usepackage{multirow}
\usepackage{placeins}
\usepackage{float}
\usepackage{pifont}
\usepackage{soul}
\usepackage[ruled,linesnumbered]{algorithm2e}
\usepackage[hidelinks]{hyperref}
\usepackage{url}
\usepackage{cleveref}
\usepackage{tikz}
\usetikzlibrary{arrows.meta,positioning}

\newlength{\meanstdwidth}
\newcommand{\meanstd}[2][0.0000]{\begingroup
  \settowidth{\meanstdwidth}{\normalfont #1}\meanstdparts#2\relax
  \endgroup
}
\def\meanstdparts#1 $\pm$ #2\relax{\makebox[\meanstdwidth][r]{#1}\ensuremath{{}\pm{}}\makebox[\meanstdwidth][l]{#2}}

\newtheorem{proposition}{Proposition}
\definecolor{myorange}{RGB}{255,165,0}

\title{Budget-Aware LLM Discovery via Cost-Calibrated Frontier Utility}

\author{
\textbf{Yansen Zhang${}^{1,2,\ast}$, Yilu Liu${}^{2}$, Tianyu Liu${}^{3}$, Jiamin Chen${}^{2}$, Xiaokun Zhang${}^{2}$, Kai Xie${}^{1}$,} \\
\textbf{Qingfu Zhang${}^{2}$, Xue Liu${}^{3}$, Yiyan Qi${}^{1,\dagger}$, Chen Ma${}^{2,\dagger}$} \\
${}^{1}$International Digital Economy Academy (IDEA) \\
${}^{2}$City University of Hong Kong \\
${}^{3}$Mohamed bin Zayed University of Artificial Intelligence (MBZUAI) \\
\texttt{yanszhang7-c@my.cityu.edu.hk}, \texttt{qiyiyan@idea.edu.cn}, \texttt{chenma@cityu.edu.hk}
}

\iclrfinalcopy
\begin{document}
\maketitle
\fancyhead{}
\renewcommand{\headrulewidth}{0pt}
\begingroup
\renewcommand{\thefootnote}{\fnsymbol{footnote}}
\footnotetext[1]{Work done as an intern at International Digital Economy Academy (IDEA).}
\footnotetext[2]{Corresponding authors: Yiyan Qi and Chen Ma.}
\endgroup

\begin{abstract}
Large language models increasingly support scientific and algorithmic discovery through inference-time search over evaluated candidates. Existing adaptive discovery controllers assign credit based only on score progress, even though prompt length, retries, and guidance calls cause search actions to incur different token costs. Under a fixed search budget, the controller must decide which frontier to develop and whether its gain justifies the realized cost before the budget is exhausted. We prove that, in the worst case, cost-blind control can forfeit all but a vanishing fraction of attainable quality as frontier count and cost heterogeneity grow. To address this limitation, we introduce \textbf{CostAda}, an adaptive controller built on \emph{cost-calibrated frontier utility}. By valuing progress relative to realized action cost and conditioning credit on the remaining budget, CostAda coordinates local exploration intensity, frontier allocation, and budgeted tactic intervention. Realized action cost and remaining budget therefore shape the search rather than serving only as accounting variables or a stopping rule. Across eight benchmarks with GLM-5 and GPT-5.4, CostAda reaches the strongest baseline's full-budget quality with at most half the budget on thirteen of sixteen benchmark--backbone pairs and achieves the strongest final quality on all sixteen pairs.
\end{abstract}

\section{Introduction}
\label{sec:introduction}

Large language models (LLMs) increasingly support scientific and algorithmic discovery through inference-time search. This process iteratively generates, evaluates, and refines candidate solutions. Across mathematical optimization, algorithm design, and executable program discovery, automated evaluators score candidates to guide further refinement \citep{romera2024mathematical,novikov2025alphaevolve,lange2025shinkaevolve}. Each search round incurs LLM costs, so solution quality depends on how a fixed search budget is allocated across competing search directions. Poor allocation can exhaust this budget before promising candidates receive enough testing and refinement.

Existing LLM discovery systems adapt their search strategies using evaluator feedback \citep{jiang2026deltaevolve,lange2025shinkaevolve,cemri2026adaevolve,liu2026evox}. However, progress-only credit assigns value to score gains without accounting for realized action cost \citep{lange2025shinkaevolve,cemri2026adaevolve}. These costs vary with prompt length, response length, and retries under token-based API pricing. A short refinement and a long-context step can therefore receive comparable credit despite consuming different amounts of the search budget. When such credit guides later choices, costly actions can keep attracting search budget without delivering greater gains.

\begin{figure*}[!t]
\centering
\resizebox{0.88\textwidth}{!}{\definecolor{utilhigh}{RGB}{0,114,178}
\definecolor{utillow}{RGB}{213,94,0}
\begin{tikzpicture}[
    font=\sffamily,
    >=Stealth,
    title/.style={font=\sffamily\bfseries\small, align=center, text=black!82},
    card/.style={draw=black!38, rounded corners=6pt, fill=white, line width=0.48pt,
        minimum width=3.35cm, minimum height=1.62cm},
    action/.style={draw=none, fill=none, align=left, inner sep=0pt,
        font=\sffamily\scriptsize, text=black!88},
    creditbase/.style={draw=orange!72!black, rounded corners=6pt, fill=orange!10,
        inner sep=5pt, minimum width=3.70cm, minimum height=1.62cm},
    creditours/.style={draw=blue!62!black, rounded corners=6pt, fill=blue!7,
        inner sep=5pt, minimum width=3.70cm, minimum height=1.62cm},
    arrow/.style={->, line width=0.56pt, draw=black!38, shorten <=1.5pt, shorten >=1.5pt},
    divider/.style={draw=black!16, line width=0.42pt, densely dashed},
    barbase/.style={draw=none, fill=orange!28, rounded corners=1.3pt},
    barA/.style={draw=none, fill=utilhigh, rounded corners=1.3pt},
    barB/.style={draw=none, fill=utillow, rounded corners=1.3pt}
]

\node[title] at (-4.25,0.98) {Traditional Control};
\node[title] at (0,0.98) {Search Actions};
\node[title] at (4.25,0.98) {CostAda Control (Ours)};

\node[creditbase] (progress) at (-4.25,-0.08) {};
\node[font=\sffamily\footnotesize\bfseries, text=orange!70!black] at (-4.25,0.32)
    {Progress-only credit};
\node[font=\sffamily\scriptsize\bfseries, text=black!75, anchor=east] at (-4.80,-0.20) {A};
\node[barbase, minimum width=1.06cm, minimum height=0.11cm, inner sep=0pt] at (-4.15,-0.20) {};
\node[font=\sffamily\scriptsize\bfseries, text=black!75, anchor=east] at (-4.80,-0.54) {B};
\node[barbase, minimum width=1.06cm, minimum height=0.11cm, inner sep=0pt] at (-4.15,-0.54) {};

\node[card] (actions) at (0,-0.08) {};
\node[action, anchor=west] at (-1.35,0.28)
    {\textbf{A:} Short refinement\\score $+0.03$ \quad cost $1\times$};
\draw[divider] (-1.45,-0.11) -- (1.50,-0.11);
\node[action, anchor=west] at (-1.35,-0.48)
    {\textbf{B:} Long-context step\\score $+0.03$ \quad cost $6\times$};

\node[creditours] (costada) at (4.25,-0.08) {};
\node[font=\sffamily\footnotesize\bfseries, text=blue!62!black, align=center] at (4.25,0.36)
    {Cost-calibrated credit};
\node[font=\sffamily\scriptsize\bfseries, text=utilhigh, anchor=east] at (3.38,-0.20) {A};
\node[barA, minimum width=0.98cm, minimum height=0.11cm, inner sep=0pt] at (4.03,-0.20) {};
\node[font=\sffamily\scriptsize\mdseries, text=utilhigh, anchor=west] at (4.68,-0.20) {higher};
\node[font=\sffamily\scriptsize\bfseries, text=utillow, anchor=east] at (3.38,-0.54) {B};
\node[barB, minimum width=0.26cm, minimum height=0.11cm, inner sep=0pt] at (3.70,-0.54) {};
\node[font=\sffamily\scriptsize\mdseries, text=utillow, anchor=west] at (4.68,-0.54) {lower};

\draw[arrow] (actions.west) -- (progress.east);
\draw[arrow] (actions.east) -- (costada.west);
\end{tikzpicture}
}
\caption{\textbf{Cost-calibrated credit for equal-gain, unequal-cost actions.} A short refinement (A) and a long-context step (B) yield the same gain ($+0.03$) at $1\times$ and $6\times$ the reference cost, respectively. Progress-only credit treats A and B equally, whereas CostAda gives A more credit than B.}
\label{fig:motivation}
\vspace{-6pt}
\end{figure*}
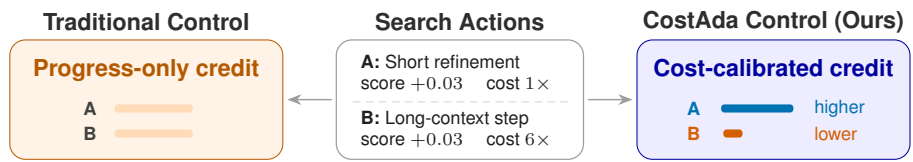

Figure~\ref{fig:motivation} illustrates this distinction with a short refinement (action A) and a long-context step (action B). Both improve the score by $0.03$, but A costs $1\times$ the reference cost and B costs $6\times$. Progress-only credit assigns equal value to A and B. At the same remaining budget, cost-calibrated credit instead gives A higher credit because it achieves the same gain at lower cost and preserves more resources for subsequent search. The value of B also depends on the remaining budget. Early in search, B may help develop a promising frontier, whereas near budget exhaustion its cost may leave too few resources to exploit that frontier. This example motivates credit that accounts for both realized action cost and the budget left for further improvements. For controllers that ignore cost, Proposition~\ref{prop:costblind} shows that the worst-case approximation ratio can vanish as frontier count and cost heterogeneity grow. Smoothing progress alone cannot resolve this cost blindness.

To address this gap, we introduce CostAda, a cost-calibrated adaptive controller for budget-aware LLM discovery. CostAda replaces progress-only credit with \emph{cost-calibrated frontier utility}, which values local frontier progress and global best-so-far improvement relative to realized action cost. Local progress indicates that a search direction is still developing, whereas global improvement measures its contribution to the strongest solution found so far. For comparable progress at the same remaining budget, a costlier action receives less credit. Early in search, CostAda gives more weight to local development because enough budget remains for a promising frontier to improve on the global best. As the budget shrinks, it shifts credit toward global improvement and penalizes costly steps more strongly, since a given call cost represents a larger fraction of the remaining budget. After each completed action, CostAda charges its realized cost and updates the corresponding frontier's credit to guide subsequent decisions using the updated budget state.

CostAda applies this credit principle through three coupled controls (Section~\ref{sec:method}). (I) Local exploration intensity favors focused refinement when recent utility is high and broader parent sampling when it is low, while narrowing exploration as the remaining budget shrinks. (II) Frontier allocation selects the next frontier by combining recent cost-calibrated global gain with an exploration bonus that decreases as the budget is spent. Local progress can guide refinement within a frontier, but only global improvement earns allocation reward. (III) Budgeted tactic intervention schedules a guide when global progress stagnates or low-yield spending accumulates, provided the remaining budget covers its estimated cost and a reserve for testing its tactics. These controls make cost and remaining budget part of search decisions, beyond their roles in accounting and stopping.

\paragraph{Contributions.} (1) We formulate LLM discovery under a fixed search budget and prove a worst-case separation between cost-blind and cost-aware control (Proposition~\ref{prop:costblind}). (2) We introduce cost-calibrated frontier utility with remaining-budget conditioning as the core credit principle. (3) We instantiate this principle in CostAda, an adaptive controller that coordinates local exploration intensity, frontier allocation, and budgeted tactic intervention. (4) Across GLM-5 and GPT-5.4, CostAda reaches the strongest baseline's full-budget quality using at most half the budget on thirteen of sixteen benchmark--backbone pairs. It also achieves the strongest final quality on all sixteen pairs.

\section{Related Work}
\label{sec:related_work}

\subsection{LLM-Guided Evolutionary Search}
LLM-guided evolutionary search uses evaluator feedback to refine candidate solutions \citep{hemberg2024evolving}. Systems such as FunSearch, AlphaEvolve, and SimpleTES apply this loop to mathematical and algorithmic discovery \citep{romera2024mathematical,novikov2025alphaevolve,ye2026evaluation}. Related methods extend it to heuristic design \citep{liu2024evolution,ye2024reevo,vanstein2025llamea}, scientific equation discovery \citep{shojaee2025llm}, and prompt optimization \citep{yang2024large,fernando2024promptbreeder,agrawal2025gepa}. Recent program-evolution frameworks study efficient sampling, candidate management, and context reuse \citep{lange2025shinkaevolve,assumpccao2025codeevolve,jiang2026deltaevolve}. TTT-Discover and PACEvolve++ further learn proposal or advisor policies during search \citep{yuksekgonul2026learning,yan2026pacevolve++}. EvoX co-evolves candidate solutions and their search strategies \citep{liu2026evox}. Progress-aware controllers such as AdaEvolve assign credit from score gains without incorporating realized action cost or the remaining budget \citep{cemri2026adaevolve}. In contrast, CostAda uses realized action cost to calibrate frontier credit and the remaining budget to adjust credit and subsequent search decisions.

\subsection{Budget-Aware LLM Systems}
LLM reasoning and agent systems face a related budget allocation problem, as each reasoning step or tool call consumes resources needed for subsequent decisions. Classical bandits with knapsacks formalize such allocation as cumulative reward maximization under resource constraints \citep{badanidiyuru2013bandits}. Token-budget-aware reasoning methods control reasoning length or adapt token allocation to query difficulty \citep{han2025token,li2025selfbudgeter,wen2025budgetthinker}. Agent methods expose remaining resources to guide tool use or jointly select models and budgets for multi-stage tasks \citep{liu2025budget,wang2025agenttts}. BG-MCTS conditions tree selection and expansion on the remaining output-token budget \citep{miyamoto2026aligning}. These reasoning and agent methods focus on completing tasks within budget, rather than continually improving candidate programs for scientific discovery. Recent work addresses budget allocation in evolutionary discovery \citep{xing2026compute,yang2026turboevolve}, but does not calibrate progress feedback by realized action cost. However, CostAda introduces cost-calibrated frontier utility, distinguishing equal gains by the cost of obtaining them and adjusting their credit with the remaining budget to guide subsequent search.

\section{Problem Formulation}
\label{sec:problem}

We consider LLM discovery over a discrete space $\mathcal{P}$ of executable candidates. A deterministic evaluator $F:\mathcal{P}\rightarrow\mathbb{R}$ assigns each candidate a scalar score, where higher is better. For tasks with multiple metrics or a minimization objective, $F$ is a direction-adjusted scalar proxy. At each iteration $t$, the LLM generates a candidate $p_t\in\mathcal{P}$, and the evaluator returns $f_t=F(p_t)$. Search reuses evaluated candidates to improve the best-so-far score $y_t=\max_{\tau\le t}f_\tau$.

We formulate this iterative refinement as a search over $K_t$ frontiers, or local candidate populations, at iteration $t$. For frontier $k\in\{1,\dots,K_t\}$, the retained candidates form a local archive $D_t^{(k)}\subseteq\mathcal{P}$ with local best score $b_t^{(k)}=\max_{p\in D_t^{(k)}}F(p)$. Across all frontiers, $y_t$ is the global best score found so far. A context builder combines a parent sampled from the selected frontier $k_t$ with archived context candidates to form the next prompt. After evaluation, the selected archive is updated, and a search controller uses the score to select subsequent search actions.

\subsection{Search Budget and Objective}

To account for differences in generation and guidance costs, we adopt a search budget $B>0$ for search-side LLM expenditure \citep{badanidiyuru2013bandits,yang2026bamas}. The realized cost $c_t$ at iteration $t$ covers candidate generation, retries, and guidance calls, excluding deterministic evaluator execution. The controller accumulates these costs as $C_t=\sum_{s\le t}c_s$ and represents the unspent budget by the remaining-budget ratio
\[
\rho_t=\max(0,1-C_t/B)\in[0,1].
\]
For credit calibration, the controller uses the normalized step cost $\tilde{c}_t=c_t/(\bar{c}+\epsilon_c)$, where $\bar{c}$ is a fixed reference generation cost shared across benchmarks and $\epsilon_c>0$ is a stabilizer. The budget constraint applies to actual cumulative expenditure $C_t$, with accounting details in Appendix~\ref{app:experimental_accounting}.

The budgeted discovery objective is to maximize expected best-so-far quality before the search-side budget is exhausted:
\[
\max_{\pi}\ \mathbb{E}[y_{\tau_B}],
\qquad
\tau_B=\max\{t:C_t\le B\},
\]
where $\pi$ is the search policy and $\tau_B$ is the last iteration whose cumulative cost stays within budget $B$. The final output is the highest-scoring candidate found by $\tau_B$.

\subsection{Failure of Cost-Blind Credit}

Call a controller \emph{cost-blind} if each action distribution depends only on action and score history, never on realized cost. A controller whose decisions rely only on progress credit and use realized cost nowhere else belongs to this class. Smoothing or combining progress signals preserves this cost-blindness. The following proposition isolates its effect when actions have different costs.

\begin{proposition}[Worst-case separation between cost-blind and cost-aware control]
\label{prop:costblind}
Consider search over $K\ge2$ frontiers with initial best score zero. Every selection increases the global best score by the same positive amount. One frontier costs $c>0$ per selection, while the others cost $\kappa c$, where $\kappa>1$ is the cost ratio. For any cost-blind controller and budget $B\ge c$, some fixed placement of the cheap frontier limits the controller's expected budgeted value to at most $\left(\tfrac{1}{K}+\tfrac{1}{\kappa}\right)\tfrac{B/c}{\lfloor B/c\rfloor}\,\mathrm{OPT}$. Here, $\mathrm{OPT}$ is the optimal value attainable within budget $B$ on that instance. A cost-aware policy that samples each frontier once and then selects by gain per realized cost attains $(1-o(1))\,\mathrm{OPT}$ on every placement as $B/c\to\infty$, with $K,\kappa$ fixed.
\end{proposition}

The proof is in Appendix~\ref{app:costblind_proof}. For fixed $K$ and $\kappa$, the finite-budget rounding factor tends to one as $B/c$ grows. The limiting upper bound $1/K+1/\kappa$ tends to zero as the frontier count $K$ and the cost ratio $\kappa$ increase. This separation motivates the cost-calibrated utility in Section~\ref{sec:method}, which restores cost-dependent credit and conditions subsequent decisions on the remaining budget.

\section{Method}
\label{sec:method}

\subsection{Overview}

\begin{figure*}[!t]
\centering
\resizebox{\textwidth}{!}{\definecolor{ctrlI}{RGB}{0,158,115}
\definecolor{ctrlII}{RGB}{213,94,0}
\definecolor{ctrlIII}{RGB}{204,121,167}
\begin{tikzpicture}[
    font=\sffamily,
    line cap=round, line join=round,
    >={Stealth[length=1.8mm, width=1.45mm, inset=0.25mm, round]},
    module/.style={draw=black!38, fill=white, rounded corners=6.5pt, line width=0.55pt,
        align=center, inner sep=4pt, minimum width=1.92cm, minimum height=0.74cm,
        font=\sffamily\scriptsize\mdseries, text=black!82},
    evalmodule/.style={module, minimum width=2.18cm},
    controller/.style={draw=blue!62!black, fill=blue!7, rounded corners=10pt,
        line width=0.65pt, minimum width=6.05cm, minimum height=1.795cm},
    pillgreen/.style={draw=ctrlI!70!black!80, fill=ctrlI!11, rounded corners=6.6pt, line width=0.55pt,
        minimum width=1.58cm, minimum height=0.72cm, align=center,
        inner xsep=3.5pt, inner ysep=3.5pt, font=\sffamily\fontsize{6.5}{7.5}\selectfont\bfseries, text=ctrlI!68!black},
    pillred/.style={draw=ctrlII!70!black!80, fill=ctrlII!9, rounded corners=6.6pt, line width=0.55pt,
        minimum width=1.58cm, minimum height=0.72cm, align=center,
        inner xsep=3.5pt, inner ysep=3.5pt, font=\sffamily\fontsize{6.5}{7.5}\selectfont\bfseries, text=ctrlII!78!black},
    pillpurple/.style={draw=ctrlIII!65!black!80, fill=ctrlIII!12, rounded corners=6.6pt, line width=0.55pt,
        minimum width=1.58cm, minimum height=0.72cm, align=center,
        inner xsep=3.5pt, inner ysep=3.5pt, font=\sffamily\fontsize{6.5}{7.5}\selectfont\bfseries, text=ctrlIII!58!black},
    state/.style={draw=black!45, fill=white, rounded corners=8pt, line width=0.55pt,
        align=center, inner xsep=7pt, inner ysep=5pt, minimum width=3.60cm, minimum height=0.90cm,
        font=\sffamily\scriptsize\mdseries, text=black!75},
    mainarrow/.style={->, draw=black!38, line width=0.65pt},
    returnarrow/.style={mainarrow, draw=black!38, line width=0.55pt},
    signal/.style={->, draw=blue!62!black, line width=0.55pt, shorten <=1.5pt, shorten >=1.5pt},
    greenarrow/.style={->, draw=ctrlI!75!black!85, line width=0.65pt, shorten <=1.5pt, shorten >=1.5pt},
    redarrow/.style={->, draw=ctrlII!80!black!85, line width=0.65pt, shorten <=1.5pt, shorten >=1.5pt},
    magentaarrow/.style={->, draw=ctrlIII!62!black!85, line width=0.65pt, shorten <=1.5pt, shorten >=1.5pt},
    label/.style={font=\sffamily\fontsize{6}{7}\selectfont\itshape, text=black!52, fill=none, inner sep=1.2pt},
    bluelabel/.style={font=\sffamily\fontsize{6}{7}\selectfont\itshape, text=blue!62!black, fill=none, inner sep=1.2pt},
    midlabel/.style={bluelabel, fill=white, inner xsep=2pt, inner ysep=1pt},
    tag/.style={circle, draw=black!14, fill=white, minimum size=0.26cm,
        inner sep=1pt, font=\sffamily\fontsize{6}{7}\selectfont\bfseries}
]

\def\costadaouterinset{0.285cm}
\def\costadallmportinset{8pt}

\node[module] (archive) at (-5.45,1.28) {Frontier\\Archive};
\node[module] (context) at (-2.42,1.28) {Context\\Builder};
\node[module] (llm) at (0.64,1.28) {Search-side\\LLM};
\node[evalmodule] (eval) at (4.05,1.28) {Deterministic\\Evaluator};
\coordinate (llmGuideIn) at ([xshift=\costadallmportinset]llm.south west);
\coordinate (llmCostOut) at ([xshift=-\costadallmportinset]llm.south east);

\draw[mainarrow] (archive.east) -- node[label, above=3pt] {sample} (context.west);
\draw[mainarrow] (context.east) -- node[label, above=3pt] {prompt} (llm.west);
\draw[mainarrow] (llm.east) -- node[label, above=3pt] {candidate} (eval.west);
\draw[returnarrow, rounded corners=8pt] (eval.north) -- (4.05,2.28) -- node[label, above=2pt] {archive update} (-5.45,2.28) -- (archive.north);

\coordinate (controlrow) at (0,-0.7875);
\node[controller, anchor=west] (ctrl) at ([xshift=\costadaouterinset]archive.west |- controlrow) {};
\node[pillred] (route) at (-4.84,-0.43) {II Frontier\\{\mdseries allocation}};
\node[pillgreen] (local) at (-3.10,-0.43) {I Local\\{\mdseries intensity}};
\node[pillpurple] (guide) at (-1.36,-0.43) {III Tactic\\{\mdseries intervention}};
\node[font=\sffamily\bfseries\scriptsize, text=black!86] at (-3.10,-1.065)
    {CostAda Controller};
\node[font=\sffamily\fontsize{6.5}{7.5}\selectfont, text=black!75] at (-3.10,-1.405)
    {Choose the next search actions};

\node[state, anchor=east] (credit) at ([xshift=-\costadaouterinset]eval.east |- ctrl.center)
    {{\bfseries\color{black!82}Budget and Credit Update}\\{\mdseries\fontsize{6.5}{8}\selectfont Remaining-budget ratio $\rho_t$}\\{\mdseries\fontsize{6.5}{8}\selectfont Cost-calibrated frontier utility $H_t$}\\{\mdseries\fontsize{6.5}{8}\selectfont Allocation reward $R_t$}};

\coordinate (costIn) at ([xshift=-0.72cm]credit.north);
\coordinate (scoreIn) at ([xshift=0.72cm]credit.north);
\draw[signal] (llmCostOut) to[out=-65,in=112]
    node[midlabel, pos=0.50] {$c_t$ cost} (costIn);
\draw[signal] (eval.south) to[out=-100,in=80]
    node[midlabel, pos=0.50] {$f_t$ score} (scoreIn);
\draw[signal] (credit.west) -- node[bluelabel, above=3pt,
    font=\sffamily\fontsize{6}{7}\selectfont\itshape] {state} (ctrl.east);

\draw[redarrow] (route.north) to[out=102,in=-78]
    node[tag, text=ctrlII!80!black, pos=0.50] {II} (archive.south);
\draw[greenarrow] (local.north) to[out=98,in=-82]
    node[tag, text=ctrlI!72!black, pos=0.50] {I} (context.south);
\draw[magentaarrow] (guide.north) to[out=78,in=-118]
    node[tag, text=ctrlIII!60!black, pos=0.50] {III} (llmGuideIn);
\end{tikzpicture}
}
\caption{\textbf{The overall architecture of CostAda.} Gains over stored best scores are calibrated by realized cost and remaining budget. The lower-right block maintains the remaining-budget ratio $\rho_t$ and smoothed credit signals $H_t$ (frontier utility) and $R_t$ (allocation reward). This state then guides local exploration intensity (I), frontier allocation (II), and budgeted tactic intervention (III).}
\label{fig:costada_overview}
\vspace{-6pt}
\end{figure*}
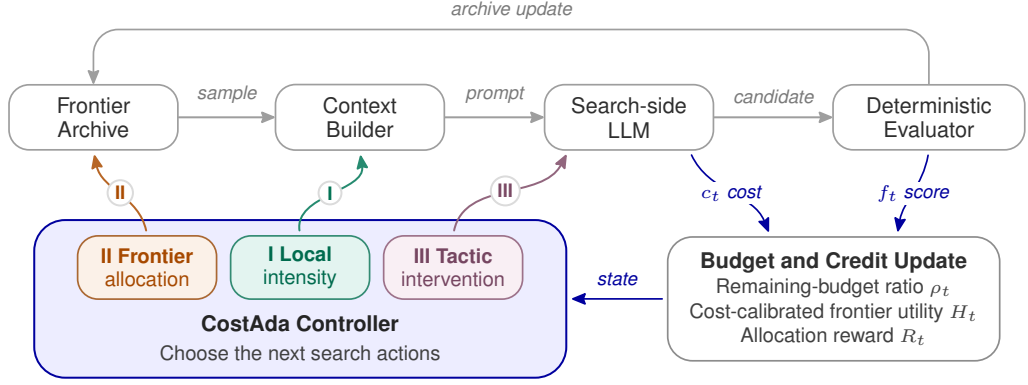

As shown in Figure~\ref{fig:costada_overview}, the CostAda controller combines cost-calibrated frontier utility with remaining-budget conditioning. Each frontier maintains a group of candidate programs in a local archive, and all frontiers share the search budget. The upper loop constructs a prompt from a selected candidate and archived context, generates and evaluates a new candidate, then updates the archive. The resulting score and realized action cost feed the lower-right block, which measures gains against previous local and global bests and updates the remaining budget and cost-calibrated credit. The controller in the lower-left block uses this updated state to choose the next search actions through three controls. Local exploration intensity responds to frontier utility, frontier allocation uses global gain relative to cost, and budgeted tactic intervention combines stagnation or low-yield spending with affordability. Decisions use the budget and credit available before each action, whose realized cost updates the state only after completion and thus informs later decisions.

\subsection{Cost-Calibrated Frontier Utility}

To assess a frontier's progress, we distinguish improvements within that frontier from improvements in the best solution found across the search. Because a candidate can improve its frontier's best score without exceeding the global best, local gain captures progress not reflected in global gain. At iteration $t$, the new candidate from frontier $k_t$ receives score $f_t$. Relative to the pre-iteration local best $b_{t-1}^{(k_t)}$ and global best $y_{t-1}$, its local gain $\delta_t^{(k_t)}$ and global gain $g_t$ are
\[
\delta_t^{(k_t)}=\max\left(\frac{f_t-b_{t-1}^{(k_t)}}{\max(|f_t|,|b_{t-1}^{(k_t)}|,1)},0\right),
\qquad
g_t=\max\left(\frac{f_t-y_{t-1}}{\max(|f_t|,|y_{t-1}|,1)},0\right).
\]
Here, both gains quantify positive improvements normalized by the larger magnitude of the candidate and reference scores. A denominator floor of one prevents amplification near zero.

These gains are weighted by the budget remaining after iteration $t$. Specifically, we set the global weight to $\lambda_t=1-\rho_t$, so that credit shifts from local development toward global improvement as the budget shrinks. We further account for realized action cost by dividing this weighted progress by the cost-calibration factor $d_t=1+\lambda_t^c\phi_t$. Here, $\phi_t=\log(1+\tilde{c}_t)$ logarithmically scales the normalized step cost from Section~\ref{sec:problem}, limiting the influence of unusually expensive calls. The corresponding cost weight $\lambda_t^c=\max(\lambda_t,\lambda_{\min})$ allows the penalty to strengthen as the budget is spent, while its fixed positive floor $\lambda_{\min}$ keeps it active from the start. Thus, the cost-calibrated frontier utility is given by
\begin{equation}
u_t^{(k_t)}=\frac{\lambda_t g_t + (1-\lambda_t)\delta_t^{(k_t)}}{d_t}.
\label{eq:utility}
\end{equation}

A step without improvement has zero utility even when the frontier has recently made progress. We therefore guide subsequent local exploration with an exponential moving average that retains recent credit while gradually discounting older observations:
\[
H_t^{(k_t)}=\alpha H_{t-1}^{(k_t)}+(1-\alpha)u_t^{(k_t)},
\]
where $\alpha\in(0,1)$ is the smoothing coefficient, with larger values placing more weight on past utility and smaller values on the latest utility. Only the selected frontier receives a new utility observation, so $H_t^{(k)}=H_{t-1}^{(k)}$ for $k\ne k_t$. When the observed utility is zero, the selected frontier's history is discounted by a factor of $\alpha$, while the incurred cost reduces the remaining budget.

\subsection{Budget-Aware Control}
\label{subsec:budget_aware_control}

The controller uses progress and cost feedback together with the remaining budget to guide subsequent search through three controls. Local exploration intensity adjusts the breadth of sampling within a frontier, frontier allocation selects the next frontier, and budgeted tactic intervention determines when to purchase guidance that the remaining budget can cover.

\paragraph{Local exploration intensity.}
Local control adjusts sampling between strong candidates and more novel candidates within the selected frontier. Initial sampling gathers feedback when a frontier is first selected. For a previously selected frontier, the default sampling rule sets intensity at iteration $t$ using its utility state and the remaining budget from iteration $t-1$:
\[
I_t^{(k_t)}=I_{\min}+(I_{\max}-I_{\min})\cdot\frac{\rho_{t-1}}{1+\sqrt{H_{t-1}^{(k_t)}+\epsilon_H}} ,
\]
where $I_{\min}$ and $I_{\max}$ are fixed lower and upper bounds on local sampling intensity and $\epsilon_H>0$ is a small stabilizer. High recent utility reflects cost-efficient progress and favors lower-intensity sampling around strong candidates. In contrast, lower recent utility raises intensity at a given remaining-budget ratio, broadening parent sampling toward novel candidates. Meanwhile, the multiplier $\rho_{t-1}$ narrows exploration as the remaining budget shrinks.

We then map this intensity to probabilities over exploration, exploitation, and balanced sampling using the fixed sampler $\pi_{\mathrm{local}}$. The resulting mode $a_t\sim\pi_{\mathrm{local}}(\cdot\mid I_t^{(k_t)})$ sets the default parent distribution and prompt guidance. Across these modes, we request the same number of local and global context programs, with sampling and retrieval rules detailed in Appendix~\ref{app:local_exploration_control}.

\paragraph{Frontier allocation.}
Frontier allocation chooses which frontier receives the next search step. Its reward credits only improvements in the global best, which advance the overall discovery objective. After iteration $t$, we apply the same cost calibration to define its reward:
\[
r_t^{(k_t)}=\frac{g_t}{d_t}.
\]
To compare recent contributions across frontiers, CostAda smooths this reward as
\[
R_t^{(k_t)}=\gamma R_{t-1}^{(k_t)}+(1-\gamma)r_t^{(k_t)},
\]
where $\gamma\in(0,1)$ is a fixed smoothing coefficient. When guidance yields substantial global improvement, a consolidation window selects the frontier on which the guide was generated, uses exploitation mode, and suppresses new interventions. Outside this window, CostAda selects the next frontier using an upper-confidence allocation rule \citep{auer2002finite}:
\[
k_{t+1}=\arg\max_k\left[R_t^{(k)}+\rho_t\,c_{\mathrm{ucb}}\sqrt{\tfrac{\log(N_t+2)}{n_t^{(k)}+1}}\right],
\]
where $c_{\mathrm{ucb}}$ is a fixed optimism scale, $n_t^{(k)}$ counts completed selections of frontier $k$, and $N_t=\sum_k n_t^{(k)}$ is the total number of completed frontier selections. The first term exploits frontiers with recent global gain relative to cost. The second gives under-sampled frontiers an exploration bonus that decreases with the remaining budget. As less budget remains to develop under-sampled frontiers, allocation increasingly favors observed global gain relative to cost.

\paragraph{Budgeted tactic intervention.}
When candidate refinement makes little global progress, a guide call can generate a batch of tactics. Because testing these tactics requires further candidate generation, we use the affordability indicator $A_t$ to check that the remaining budget covers the estimated guide cost and a testing reserve. The stagnation indicator $P_t$ equals one after a sustained absence of meaningful global progress, with patience adapting to the remaining budget and frontier count. In addition, $Y_t$ detects accumulated low-yield spending, allowing intervention before the stagnation requirement is met. After iteration $t$, we schedule guidance for the next eligible iteration using
\[
\mathcal{I}_t=A_t\land(P_t\lor Y_t),
\]
with $\mathcal{I}_t=0$ during consolidation, an active tactic batch, or a pause after unsuccessful guidance.

The candidate generations that test a guide's tactic batch form a guide cycle. Within this cycle, the guide call and all candidate generations that test its tactics are charged to the same search budget. Before guidance produces substantial global improvement, CostAda favors breakthrough tactics that propose new solution strategies. After successful guidance, refinement becomes the default mode for later guides, focusing on improvements to existing solutions. If the tactic batch is exhausted without substantial improvement, we pause new guide calls for $K_t$ further iterations while search continues. Further details can be found in Appendix~\ref{app:budgeted_tactic_intervention}.

\subsection{Overall Procedure}

In summary, each iteration selects a frontier and sampling mode from the remaining budget and stored credit, with any active consolidation window taking precedence. Any scheduled guide is generated before the candidate, whose prompt combines the selected parent program, archived context, and the active tactic when available. After evaluation, all search-side calls are charged, and gains are computed against the pre-step best scores before the archives and credit are updated. The resulting evidence determines whether to schedule an intervention for the next eligible iteration. The loop stops after reaching the budget threshold or iteration cap and returns the best candidate found. For reproducibility, Appendix~\ref{app:costada_controller_details} provides the pseudocode (Algorithm~\ref{alg:costada}), supplementary control rules, and controller settings shared across all eight benchmarks.

\section{Experiments}
\label{sec:experiments}

\subsection{Experimental Setup}
\label{sec:experimental-setting}

\paragraph{Benchmarks.}
Following prior LLM discovery studies \citep{liu2026skydiscover,cemri2026adaevolve,liu2026evox}, we evaluate on eight benchmarks spanning geometric packing, extremal geometry, additive combinatorics, and executable program optimization. Together, these benchmarks cover diverse constraints and optimization objectives in iterative program discovery. Further details, including benchmark definitions and score directions, are provided in Appendix~\ref{app:benchmark_details}.

\paragraph{Baselines.}
We compare CostAda with AdaEvolve \citep{cemri2026adaevolve} and EvoX \citep{liu2026evox}, two recent leading methods for adaptive LLM discovery. AdaEvolve is the closest progress-aware controller, adapting local exploration, frontier allocation, and intervention from score improvements. EvoX jointly evolves candidates and their search strategies, providing a complementary approach to adapting discovery. Neither baseline conditions decisions on remaining budget. Within each setting, all methods share the evaluator, search scaffold, backbone, fixed token prices, and budget protocol. This shared setting focuses the comparison on search-control mechanisms.

\paragraph{Backbones and budgets.}
We use GLM-5 \citep{glm5team2026glm5vibecodingagentic} and GPT-5.4 \citep{openai2026gpt54} with respective dollar budgets $B{=}1$ and $B{=}5$. Appendix~\ref{app:experimental_accounting} provides accounting details and fixed token prices. Each run stops after the first completed iteration with cumulative cost at or above $B$, or after 100 iterations (budget diagnostics in Appendix~\ref{app:budget_adherence}). To track search progress, we record results from the same runs at the budget cutoffs $0.25B$, $0.5B$, $0.75B$, and $B$, using the first completed iteration reaching or exceeding each threshold.

\paragraph{Metrics.}
Following the same studies \citep{liu2026skydiscover,cemri2026adaevolve,liu2026evox}, we measure final quality with the native benchmark objective of the best-so-far candidate, which we record at budget cutoffs rather than iteration counts (\textsc{BestScore@Budget}). Table~\ref{tab:costada-math8-final-quality} lists the native units and score directions. Because these objectives are not comparable across benchmarks, we also record a direction-adjusted normalized score. To complement final quality, we compute normalized \textsc{AUC} to summarize the best-so-far normalized score over the budget path and capture how early strong solutions are found (Appendix~\ref{app:normalized_results}). Across three independent runs per setting, as in AdaEvolve and EvoX \citep{cemri2026adaevolve,liu2026evox}, we report the mean and sample standard deviation for both metrics, together with the strongest final objective (Best).

\begin{table}[!t]
\caption{
\textbf{Solution quality under equal budgets within each backbone.}
Arrows indicate the preferred direction. The strongest displayed values, including ties, are bolded.
We adopt the human reference values reported in prior work \citep{novikov2025alphaevolve,cemri2026adaevolve,liu2026evox}.
}
\label{tab:costada-math8-final-quality}
\begin{center}
\begin{adjustbox}{max width=\linewidth}
\begin{tabular}{l c cc cc cc}
\toprule
\multirow{2}{*}{\textbf{Benchmark}}
& \multirow{2}{*}{\textbf{Human}}
& \multicolumn{2}{c}{\textbf{AdaEvolve}}
& \multicolumn{2}{c}{\textbf{EvoX}}
& \multicolumn{2}{c}{\textbf{CostAda}} \\
\cmidrule(lr){3-4}\cmidrule(lr){5-6}\cmidrule(lr){7-8}
& & \textbf{Mean $\pm$ Std} & \textbf{Best}
& \textbf{Mean $\pm$ Std} & \textbf{Best}
& \textbf{Mean $\pm$ Std} & \textbf{Best} \\
\midrule
\multicolumn{8}{l}{\emph{Backbone: GLM-5}} \\
Circle Packing $\uparrow$ & 2.6340 & \meanstd{2.5617 $\pm$ 0.0773} & 2.6224 & \meanstd{2.5717 $\pm$ 0.0752} & 2.6181 & \textbf{\meanstd{2.6306 $\pm$ 0.0054}} & \textbf{2.6360} \\
Circle Packing Rect $\uparrow$ & 2.3640 & \meanstd{2.2089 $\pm$ 0.1635} & 2.3495 & \meanstd{2.1989 $\pm$ 0.2563} & 2.3540 & \textbf{\meanstd{2.3525 $\pm$ 0.0081}} & \textbf{2.3572} \\
Heilbronn Convex $\uparrow$ & 0.0306 & \meanstd{0.0229 $\pm$ 0.0015} & 0.0247 & \meanstd{0.0235 $\pm$ 0.0012} & 0.0245 & \textbf{\meanstd{0.0263 $\pm$ 0.0018}} & \textbf{0.0280} \\
Heilbronn Triangle $\uparrow$ & 0.0360 & \meanstd{0.0303 $\pm$ 0.0011} & 0.0314 & \meanstd{0.0220 $\pm$ 0.0050} & 0.0257 & \textbf{\meanstd{0.0314 $\pm$ 0.0003}} & \textbf{0.0317} \\
MinMaxDist $(n{=}16,d{=}2)$ $\downarrow$ & 12.89 & \meanstd{14.06 $\pm$ 1.30} & 13.29 & \meanstd{14.17 $\pm$ 1.92} & 13.06 & \textbf{\meanstd{13.11 $\pm$ 0.29}} & \textbf{12.89} \\
MinMaxDist $(n{=}14,d{=}3)$ $\downarrow$ & \hphantom{0}4.17 & \meanstd{4.38 $\pm$ 0.08} & \hphantom{0}4.31 & \meanstd{4.20 $\pm$ 0.03} & \textbf{\hphantom{0}4.17} & \textbf{\meanstd{4.17 $\pm$ 0.00}} & \textbf{\hphantom{0}4.17} \\
Third Autocorrelation $\downarrow$ & 1.4581 & \meanstd{1.4685 $\pm$ 0.0014} & 1.4674 & \meanstd{1.5481 $\pm$ 0.1308} & 1.4720 & \textbf{\meanstd{1.4658 $\pm$ 0.0037}} & \textbf{1.4620} \\
Signal Processing $\uparrow$ & -- & \meanstd{0.6393 $\pm$ 0.0240} & 0.6656 & \meanstd{0.6502 $\pm$ 0.0430} & 0.6927 & \textbf{\meanstd{0.7222 $\pm$ 0.0127}} & \textbf{0.7364} \\
\midrule
\multicolumn{8}{l}{\emph{Backbone: GPT-5.4}} \\
Circle Packing $\uparrow$ & 2.6340 & \meanstd{2.6196 $\pm$ 0.0030} & 2.6216 & \meanstd{2.5851 $\pm$ 0.0510} & 2.6215 & \textbf{\meanstd{2.6248 $\pm$ 0.0046}} & \textbf{2.6280} \\
Circle Packing Rect $\uparrow$ & 2.3640 & \meanstd{2.3148 $\pm$ 0.0357} & 2.3501 & \meanstd{2.3474 $\pm$ 0.0011} & 2.3487 & \textbf{\meanstd{2.3577 $\pm$ 0.0050}} & \textbf{2.3621} \\
Heilbronn Convex $\uparrow$ & 0.0306 & \meanstd{0.0197 $\pm$ 0.0020} & 0.0220 & \meanstd{0.0223 $\pm$ 0.0031} & \textbf{0.0252} & \textbf{\meanstd{0.0244 $\pm$ 0.0008}} & \textbf{0.0252} \\
Heilbronn Triangle $\uparrow$ & 0.0360 & \meanstd{0.0307 $\pm$ 0.0008} & 0.0314 & \meanstd{0.0263 $\pm$ 0.0035} & 0.0304 & \textbf{\meanstd{0.0330 $\pm$ 0.0009}} & \textbf{0.0337} \\
MinMaxDist $(n{=}16,d{=}2)$ $\downarrow$ & 12.89 & \meanstd{13.00 $\pm$ 0.00} & 13.00 & \meanstd{13.00 $\pm$ 0.00} & 13.00 & \textbf{\meanstd{12.93 $\pm$ 0.06}} & \textbf{12.89} \\
MinMaxDist $(n{=}14,d{=}3)$ $\downarrow$ & \hphantom{0}4.17 & \meanstd{4.55 $\pm$ 0.15} & \hphantom{0}4.46 & \meanstd{4.44 $\pm$ 0.02} & \hphantom{0}4.42 & \textbf{\meanstd{4.42 $\pm$ 0.08}} & \textbf{\hphantom{0}4.32} \\
Third Autocorrelation $\downarrow$ & 1.4581 & \meanstd{1.5081 $\pm$ 0.0069} & 1.5012 & \meanstd{1.4824 $\pm$ 0.0042} & 1.4788 & \textbf{\meanstd{1.4717 $\pm$ 0.0074}} & \textbf{1.4631} \\
Signal Processing $\uparrow$ & -- & \meanstd{0.7182 $\pm$ 0.0137} & 0.7334 & \meanstd{0.6767 $\pm$ 0.0179} & 0.6968 & \textbf{\meanstd{0.7672 $\pm$ 0.0292}} & \textbf{0.7990} \\
\bottomrule
\end{tabular}
\end{adjustbox}
\end{center}
\vspace{-10pt}
\end{table}

\subsection{Solution Quality Under Equal Budgets}
\label{subsec:final_quality}

As shown in Table~\ref{tab:costada-math8-final-quality}, we make the following observations about final quality under equal budgets. First, CostAda's advantage over both adaptive baselines is consistent across benchmarks and backbones. It achieves the strongest mean final objective across all eight benchmarks with GLM-5 and GPT-5.4, and its Best is also strongest or tied in every setting. Moreover, CostAda's mean matches or exceeds each baseline's Best on six of the eight benchmarks under GLM-5 and seven under \mbox{GPT-5.4}, reflecting consistently strong runs. Second, CostAda improves program quality even when the strongest baseline changes with the backbone. On Signal Processing, it raises the mean final objective by $11.1\%$ over EvoX under GLM-5 and by $6.8\%$ over AdaEvolve under GPT-5.4. The ablations in Appendix~\ref{app:component_ablations} attribute much of this gain to cost calibration and remaining-budget conditioning. Removing remaining-budget conditioning under GLM-5, or either component under GPT-5.4, lowers the final Signal Processing objective below that of the strongest baseline.

Finally, CostAda matches the reported human references on both MinMaxDist tasks and exceeds the reference on Circle Packing. Under GLM-5, CostAda's runs on MinMaxDist $(n{=}14,d{=}3)$ all reach the reference value of $4.17$, and its Best matches the reference on MinMaxDist $(n{=}16,d{=}2)$ under both backbones. On the latter task under GPT-5.4, AdaEvolve and EvoX remain at $13.00$ in every run, whereas CostAda lowers the mean to $12.93$. On Circle Packing under GLM-5, CostAda's Best exceeds the reference ($2.6360$ vs.\ $2.6340$), and its standard deviation is about one fourteenth of those of AdaEvolve and EvoX. Because each baseline's Best comes within $0.013$ of CostAda's mean, its mean advantage on this task stems mainly from consistently avoiding weak runs.

\subsection{Budget Efficiency}
\label{subsec:budget_cutoffs}

\begin{figure}[!t]
\centering
\includegraphics[width=0.8\linewidth]{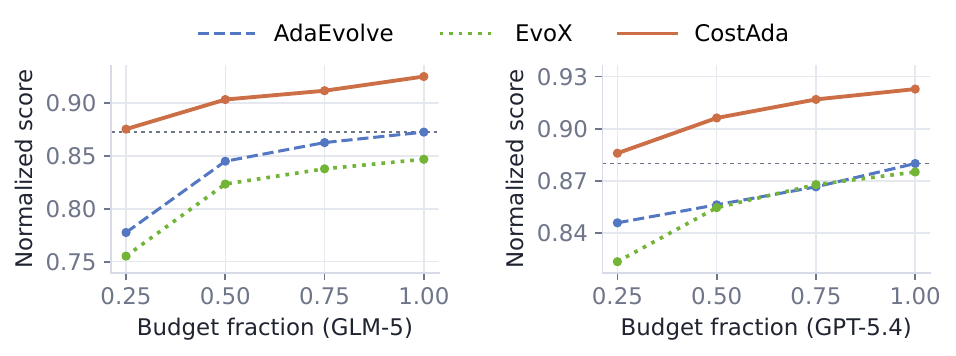}
\vspace{-6pt}
\caption{
\textbf{Mean normalized score at different budget fractions.} At each cutoff, scores are averaged over eight benchmarks under GLM-5 (left) and GPT-5.4 (right). The panels use independent vertical scales. Dotted lines mark the strongest baseline's full-budget score in each panel. Per-benchmark normalized scores and \textsc{AUC} values appear in Appendix~\ref{app:normalized_results}.
}
\label{fig:budget-path-summary}
\vspace{-8pt}
\end{figure}

Figure~\ref{fig:budget-path-summary} reveals that CostAda finds strong solutions early and stays ahead as spending increases. First, CostAda needs substantially less budget to reach the strongest baseline's final quality. At a quarter of the budget, its mean normalized score under both backbones already exceeds the strongest baseline's full-budget score, marked by the dotted line. Per benchmark, it reaches the strongest baseline's full-budget quality with at most half the budget on thirteen of sixteen benchmark--backbone pairs (Appendix~\ref{app:additional_budget_cutoffs}). Cost calibration and remaining-budget conditioning both contribute to this early advantage, as the ablations in Section~\ref{subsec:component_ablations} show. Second, the lead persists along the entire budget path. CostAda's curve lies at least $0.04$ above both baselines at each of the four cutoffs, and it achieves the highest normalized \textsc{AUC} on fifteen of sixteen pairs (Appendix~\ref{app:normalized_results}).

Finally, the early advantage also holds when a smaller budget is specified before search. We configure CostAda directly with $B{=}0.5$ under GLM-5 and compare methods at the same absolute costs (Appendix~\ref{app:small_budget_rerun_vs_cutoff}). These runs outperform the strongest baseline result in all 16 benchmark--cutoff cells. They also match or improve on CostAda's corresponding cutoffs from runs configured with $B{=}1$ in 14 cells. Here, the smaller budget conditions credit and control from the start of search.

\subsection{Ablation Study}
\label{subsec:component_ablations}

We compare CostAda with three variants in Table~\ref{tab:component-ablation-auc}, each keeping all three controls and removing one budget-aware component. Specifically, w/o cost calibration sets $d_t=1$, leaving progress-only credit, and w/o budget conditioning fixes $\lambda_t=\lambda_t^c=0.5$ while dropping $\rho_t$ from local intensity and the exploration bonus. The variant w/o intervention gating schedules the same guidance without the $A_t$ and $Y_t$ gates. Each component raises normalized \textsc{AUC} in all six settings, with the largest effect varying by backbone and task. First, cost calibration matters most on Heilbronn Triangle and Signal Processing under GPT-5.4, where its removal produces the largest losses. Second, remaining-budget conditioning matters most on the same two tasks under GLM-5. On Heilbronn Triangle, removing it still yields higher final and half-budget objectives than removing cost calibration (Appendix~\ref{app:component_ablations}). The lower \textsc{AUC} of this variant ($0.4073$ vs.\ $0.4654$) therefore reflects weaker quality earlier in the search, which final quality alone would miss. Finally, intervention gating matters most on Circle Packing, where its removal lowers \textsc{AUC} by $0.1039$ under GLM-5 and $0.0423$ under GPT-5.4.

\begin{table}[!t]
\caption{\textbf{Ablation study of CostAda.} Entries report normalized \textsc{AUC} as \textbf{Mean $\pm$ Std}. Higher is better, and the best means within each benchmark and backbone are bolded.}
\label{tab:component-ablation-auc}
\begin{center}
\begin{adjustbox}{max width=\linewidth}
\begin{tabular}{l l c c c}
\toprule
\textbf{Backbone} & \textbf{Variant} & \textbf{Circle Packing} & \textbf{Heilbronn Triangle} & \textbf{Signal Processing} \\
\midrule
\multirow{4}{*}{GLM-5}
& CostAda & \textbf{\meanstd{0.9517 $\pm$ 0.0095}} & \textbf{\meanstd{0.7831 $\pm$ 0.0486}} & \textbf{\meanstd{0.6546 $\pm$ 0.0444}} \\
& w/o cost calibration & \meanstd{0.9053 $\pm$ 0.0376} & \meanstd{0.4654 $\pm$ 0.0614} & \meanstd{0.6355 $\pm$ 0.0200} \\
& w/o budget conditioning & \meanstd{0.9161 $\pm$ 0.0316} & \meanstd{0.4073 $\pm$ 0.0426} & \meanstd{0.6146 $\pm$ 0.0134} \\
& w/o intervention gating & \meanstd{0.8478 $\pm$ 0.0642} & \meanstd{0.4494 $\pm$ 0.1622} & \meanstd{0.6293 $\pm$ 0.0140} \\
\midrule
\multirow{4}{*}{GPT-5.4}
& CostAda & \textbf{\meanstd{0.9825 $\pm$ 0.0050}} & \textbf{\meanstd{0.7951 $\pm$ 0.0253}} & \textbf{\meanstd{0.7022 $\pm$ 0.0224}} \\
& w/o cost calibration & \meanstd{0.9648 $\pm$ 0.0276} & \meanstd{0.6748 $\pm$ 0.0459} & \meanstd{0.6333 $\pm$ 0.0028} \\
& w/o budget conditioning & \meanstd{0.9673 $\pm$ 0.0079} & \meanstd{0.7494 $\pm$ 0.0615} & \meanstd{0.6557 $\pm$ 0.0192} \\
& w/o intervention gating & \meanstd{0.9402 $\pm$ 0.0246} & \meanstd{0.6949 $\pm$ 0.0259} & \meanstd{0.6717 $\pm$ 0.0116} \\
\bottomrule
\end{tabular}
\end{adjustbox}
\end{center}
\vspace{-10pt}
\end{table}

\subsection{Case Study}
\label{subsec:case_study}

\begin{figure}[!t]
\centering
\includegraphics[width=0.88\linewidth]{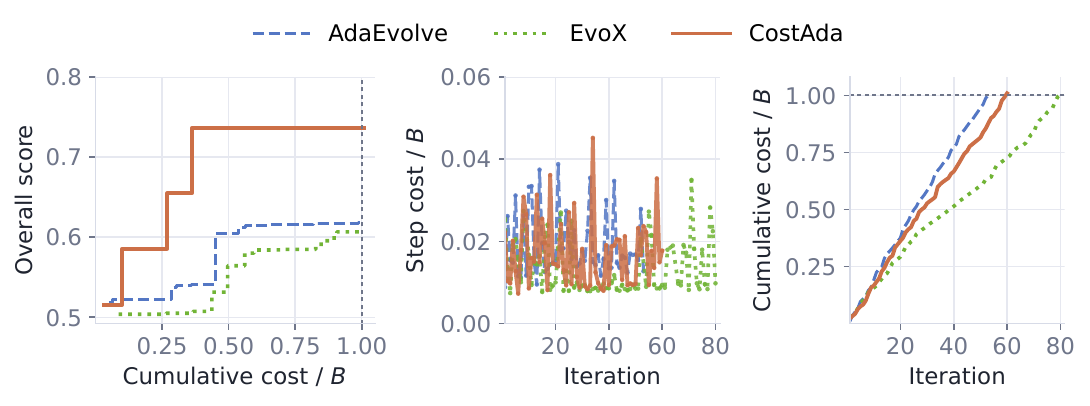}
\vspace{-6pt}
\caption{
\textbf{Solution quality and search cost for Signal Processing under GLM-5.}
Panels show the best-so-far benchmark objective versus cumulative realized cost (left), realized step cost versus iteration (middle), and cumulative realized cost versus iteration (right). Costs are normalized by the shared search budget $B$, with a dotted line at $1$.
}
\label{fig:glm5-signal-processing-budget-trajectories}
\vspace{-8pt}
\end{figure}

Figure~\ref{fig:glm5-signal-processing-budget-trajectories} follows each method's search on Signal Processing under GLM-5 to relate solution quality to realized cost. In the left panel, CostAda surpasses both baselines' final objectives within roughly a third of the budget and keeps that lead. Across runs, CostAda's half-budget mean in Appendix~\ref{app:additional_budget_cutoffs} ($0.6924$) likewise exceeds the strongest baseline's full-budget mean ($0.6502$). The middle panel shows four- to sixfold variation in step cost within each run, which cost calibration accounts for when assigning credit. In the right panel, the same budget lasts 53 iterations for AdaEvolve, 60 for CostAda, and 80 for EvoX, which illustrates why we compare methods at budget cutoffs rather than at iteration counts. The largest rise of CostAda's objective, at about $0.36B$ in the left panel, comes from the tactics of a refinement guide. Low-yield spending triggered this guide before the stagnation threshold was reached, so realized cost moved the intervention earlier (Appendix~\ref{app:worked_example}).

\section{Conclusion}
\label{sec:conclusion}

We presented CostAda, a cost-calibrated controller for budget-aware LLM discovery. Our analysis shows that the worst-case approximation ratio of cost-blind control can vanish as frontier count and cost heterogeneity grow. This limitation motivates CostAda's cost-calibrated frontier utility and remaining-budget conditioning for local exploration intensity, frontier allocation, and budgeted tactic intervention. Across eight benchmarks with GLM-5 and GPT-5.4, CostAda reaches the strongest baseline's full-budget quality with at most half the budget on thirteen of sixteen benchmark--backbone pairs. It also achieves the strongest final quality on all sixteen pairs.

\section*{AI Use Statement}

Generative AI tools assisted with method implementation and refinements to mathematical
statements and proofs. They also provided feedback on experimental comparisons and supported
result interpretation, literature retrieval, and language polishing.

\bibliographystyle{iclr2026_conference}
\bibliography{06_references}

\clearpage
\appendix

\renewcommand{\bottomfraction}{0.9}
\renewcommand{\topfraction}{0.95}
\renewcommand{\textfraction}{0.05}
\renewcommand{\floatpagefraction}{0.85}
\setcounter{topnumber}{5}
\setcounter{bottomnumber}{4}
\setcounter{totalnumber}{8}
\makeatletter
\setlength{\@fptop}{0pt}
\setlength{\@fpsep}{10pt plus 3pt}
\setlength{\@fpbot}{0pt plus 1fil}
\makeatother
\setlength{\floatsep}{12pt plus 3pt minus 3pt}
\setlength{\textfloatsep}{14pt plus 3pt minus 3pt}
\setlength{\intextsep}{12pt plus 3pt minus 3pt}

\section{Proof of Proposition~\ref{prop:costblind}}
\label{app:costblind_proof}

\paragraph{Instance.}
Fix $g>0$ and the parameters of Proposition~\ref{prop:costblind}, with initial score $y_0=0$. Every frontier selection deterministically improves the global best score by $g$. The budgeted value is therefore $g$ times the number of selections completed within budget $B$. Consider the $K$ instances indexed by the position $j$ of the cheap frontier. Frontier $j$ costs $c$ per selection, and all others cost $\kappa c$.

\paragraph{Optimum.}
Selecting the cheap frontier at every step completes $\lfloor B/c\rfloor$ selections, so $\mathrm{OPT}=g\lfloor B/c\rfloor$. No policy can complete more equal-gain selections within the same budget.

\paragraph{Upper bound for cost-blind controllers.}
Fix any possibly randomized cost-blind controller and couple the instances using the same internal randomness. Every selection returns gain $g$, so its action--score history is independent of the cheap frontier's position until termination. Extend the action sequence past budget termination solely for the counting argument, and denote this common sequence by $\sigma$. Condition on $\sigma$. For placement $j$, let $N_j$ count selections completed within budget, and let $V_j(n)$ count visits to frontier $j$ among the first $n$ selections. The first $n$ selections cost $c\left(\kappa n-(\kappa-1)V_j(n)\right)$, so the budget constraint at $n=N_j$ gives
\[
N_j\ \le\ \frac{B}{\kappa c}+\frac{\kappa-1}{\kappa}\,V_j(N_j).
\]
Every selection costs at least $c$, so $N_{\max}=\max_j N_j\le B/c$. Since each $V_j$ is nondecreasing,
\[
V_j(N_j)\le V_j(N_{\max})
\]
for every $j$, and $\sum_jV_j(N_{\max})=N_{\max}$. Summing the preceding bound over the $K$ placements and dividing by $K$ gives
\[
\frac{1}{K}\sum_{j=1}^{K}N_j
\ \le\
\frac{B}{\kappa c}+\frac{\kappa-1}{\kappa}\cdot\frac{N_{\max}}{K}
\ \le\
\left(\frac{1}{\kappa}+\frac{1}{K}\right)\frac{B}{c}.
\]
The bound holds for every sequence, so taking expectation over this internal randomness gives
\[
\frac{1}{K}\sum_{j=1}^{K}\mathbb{E}[N_j]
\ \le\
\left(\frac{1}{\kappa}+\frac{1}{K}\right)\frac{B}{c}.
\]
Hence some fixed placement $j^\star$ has expected selection count no larger than this average. This placement may depend on the controller and budget, but not on its realized randomness. Using the value $gN_{j^\star}$ and $\mathrm{OPT}=g\lfloor B/c\rfloor$ yields
\[
\mathbb{E}[gN_{j^\star}]
\ \le\
\left(\frac{1}{K}+\frac{1}{\kappa}\right)
\frac{B/c}{\lfloor B/c\rfloor}\,\mathrm{OPT}.
\]
The rounding factor tends to one as $B/c\to\infty$.

\paragraph{Achievability with cost-aware control.}
For sufficiently large $B$, select every frontier once at total cost $\left(1+(K-1)\kappa\right)c$. The realized costs identify the cheap frontier, which is then selected for the remaining budget. This policy completes at least $\lfloor B/c\rfloor-(K-1)\kappa$ selections on every placement. Ranking by gain per realized cost therefore attains at least $\mathrm{OPT}-(K-1)\kappa g=(1-o(1))\,\mathrm{OPT}$ as $B/c\to\infty$ with $K$ and $\kappa$ fixed. \hfill$\square$

\paragraph{Remarks.}
When $K,\kappa\to\infty$ with $K\kappa c/B\to0$, cost-blind control has a vanishing worst-case approximation ratio, while the constructed cost-aware policy is asymptotically optimal. The vanishing-ratio conclusion requires growing frontier count and cost ratio. Each frontier has a fixed cost, so the construction requires neither heavy-tailed costs nor nonlinear token pricing.

The construction uses equal gains deliberately. Credit based only on score progress receives no signal that separates the frontiers, whether the signal is raw, smoothed, or optimism adjusted. The failure therefore belongs to the cost-blind class rather than a particular rule. Budget termination only fixes the realized prefix length $N_j$. It cannot alter an earlier action, and no later action contributes to the budgeted value. The allocation question is related to bandits with knapsacks \citep{badanidiyuru2013bandits}, where budget-constrained optimality also favors arms with high expected gain per unit cost. The broader discovery setting includes frontier development, nonstationarity from growing archives, and interventions. The proposition isolates cost-blindness from these difficulties.

\section{CostAda Controller Details}
\label{app:costada_controller_details}

\subsection{Controller State and Overall Loop}
\label{app:controller_state_loop}

We detail how the three controls use cost-calibrated credit and the remaining budget, with their interaction summarized in Algorithm~\ref{alg:costada}. To track intervention timing, guide outcomes, and temporary control overrides, the controller maintains the evidence state
\[
E_t=(\nu_t,L_t,Z_t,o_t,w_t,\kappa_t,\omega_t,G_t,\mathcal{G}_t).
\]
Table~\ref{tab:costada-controller-state} summarizes the role of each state variable. The controller settings in Table~\ref{tab:costada-controller-constants} remain fixed across all eight benchmarks without per-task tuning.

\begin{table}[!htbp]
\caption{\textbf{CostAda evidence state.} State variables support budgeted tactic intervention and are updated from realized score and cost evidence.}
\label{tab:costada-controller-state}
\begin{center}
\begin{adjustbox}{max width=\linewidth}
\begin{tabular}{p{2.3cm} p{10.8cm}}
\toprule
\textbf{State} & \textbf{Role} \\
\midrule
$\nu_t$ & Counts consecutive steps without meaningful global progress, providing evidence for the stagnation criterion. \\
$L_t$ & Accumulates realized cost over consecutive steps with negligible utility and allocation reward to track low-yield spending. \\
$Z_t$ & Records whether guidance has produced substantial global improvement during the run. \\
$o_t$ & Stores a temporary intervention-mode correction after a failed guide cycle. \\
$w_t$ & Counts the remaining search steps in the consolidation window. \\
$\kappa_t$ & Stores the frontier selected for consolidation while $w_t>0$. \\
$\omega_t$ & Counts post-step intervention decisions that remain suppressed by guide backoff at the start of iteration $t$. \\
$G_t$ & Indicates whether the tactic batch has unused generation attempts remaining after step $t$. \\
$\mathcal{G}_t$ & Stores an open guide-credit record with its baseline score, mode, and originating frontier; empty after success or unsuccessful exhaustion. \\
\bottomrule
\end{tabular}
\end{adjustbox}
\end{center}
\end{table}

\begin{algorithm}[ht]
\DontPrintSemicolon
\SetAlgoLined
\SetKwInOut{KwIn}{Input}
\SetKwInOut{KwOut}{Output}
\newcommand{\algcomment}[1]{\textcolor{blue}{\texttt{#1}}}\SetCommentSty{algcomment}\SetKwComment{ccm}{\textcolor{blue}{\texttt{//}}~}{}

\KwIn{Initial frontiers $\{D_0^{(k)}\}_{k=1}^{K_0}$, deterministic evaluator $F$, budget $B$, reference cost $\bar{c}$, controller constants $\Theta$ (Table~\ref{tab:costada-controller-constants}), iteration cap $T_{\max}$}
\KwOut{Best-so-far candidate $p^\star$}

Initialize $p^\star$, global best $y_0$, and local bests $\{b_0^{(k)}\}$ from the initial archives; set cumulative cost $C_0\leftarrow0$, utility states $\{H_0^{(k)}\leftarrow0\}$, allocation states $\{R_0^{(k)}\leftarrow0\}$, visit counts $\{n_0^{(k)}\leftarrow0\}$, and intervention evidence states\;

\For{$t \leftarrow 1$ \KwTo $T_{\max}$}{
    \ccm{Choose search actions from the available budget and credit}
    Form the pre-step budget state $\rho_{t-1}\leftarrow\max(0,1-C_{t-1}/B)$\;
    \eIf{$w_{t-1}>0$}{
        Set $k_t\leftarrow\kappa_{t-1}$ and $a_t\leftarrow\mathrm{exploitation}$\;
    }{
        Select $k_t$ using $R_{t-1}^{(k)}$ and the $\rho_{t-1}$-scaled exploration bonus\;
        Choose $a_t$ using initial sampling or local intensity from $H_{t-1}^{(k_t)}$ and $\rho_{t-1}$\;
    }

    \ccm{Generate and evaluate the candidate}
    Execute any scheduled guide that remains available and affordable; open a credit record for a valid tactic batch, storing the baseline, mode, and originating frontier\;
    Draw parent $p_t^{\mathrm{par}}$ according to $a_t$ and retrieve context $\mathcal{C}_t$ from local and global archives; use the global best parent while a tactic batch is active\;
    Generate $p_t$ with the context and any active tactic, then evaluate $f_t$\;

    \ccm{Update the budget from realized search-side cost}
    Charge all search-side LLM calls in the step to obtain $c_t$; update $C_t$, $\rho_t$, and normalized cost $\tilde{c}_t$\;

    \ccm{Cost-calibrated credit with remaining-budget conditioning}
    Compute gains $\delta_t^{(k_t)}$ and $g_t$ against $b_{t-1}^{(k_t)}$ and $y_{t-1}$\;
    Compute $\lambda_t$, $\lambda_t^c$, and $d_t$; obtain frontier utility $u_t^{(k_t)}$ from Eq.~\eqref{eq:utility} and allocation reward $r_t^{(k_t)}=g_t/d_t$\;
    Update the selected archive $D_t^{(k_t)}$, local and global best scores, and $p^\star$\;

    Smooth $u_t^{(k_t)}$ into $H_t^{(k_t)}$ and $r_t^{(k_t)}$ into $R_t^{(k_t)}$; update visit counts and stagnation and low-yield evidence; carry forward states for unselected frontiers\;

    Evaluate guide success and exhaustion; retain the credited frontier, update $Z_t$ and mode correction $o_t$, then close any resolved credit record while retaining available tactics\;
    Update $w_t,\kappa_t$ from guide success and its originating frontier; reset $L_t\leftarrow0$ when consolidation starts\;
    Apply guide backoff after unsuccessful tactic exhaustion\;

    \ccm{Schedule budgeted tactic intervention from updated evidence}
    Compute $(\mathcal{I}_t,m_t)$ using the full intervention rule in Appendix~\ref{app:budgeted_tactic_intervention}\;
    \If{$\mathcal{I}_t=1$}{
        Schedule a budgeted guide/tactic intervention with mode $m_t$ for the next eligible iteration\;
    }
    Update the start-of-next-iteration backoff counter $\omega_{t+1}$\;
    \ccm{Budget stop}
    \If{$C_t \ge B$}{
        \textbf{break}\;
    }
}
Return $p^\star$\;
\caption{CostAda control loop}
\label{alg:costada}
\end{algorithm}

\label{app:controller_constants}
\begin{table}[!htbp]
\caption{\textbf{CostAda controller constants.} The listed values govern local exploration intensity, frontier allocation, and budgeted tactic intervention and remain fixed across all eight benchmarks.}
\label{tab:costada-controller-constants}
\begin{center}
\begin{adjustbox}{max width=\linewidth}
\begin{tabular}{@{}>{\raggedright\arraybackslash}m{1.5cm} >{\raggedright\arraybackslash}m{3.65cm} >{\centering\arraybackslash}m{1.9cm} >{\raggedright\arraybackslash}m{5.6cm}@{}}
\toprule
\textbf{Symbol} & \textbf{Role} & \textbf{Value} & \textbf{Used for} \\
\midrule
$I_{\min}, I_{\max}$ & Local intensity bounds & $0.15,\ 0.50$ & Lower and upper bounds on local sampling intensity. \\
\addlinespace[2pt]
$\alpha$ & Utility smoothing & $0.9$ & Exponential moving average that forms $H_t^{(k)}$ from utility. \\
\addlinespace[2pt]
$\gamma$ & Allocation smoothing & $0.9$ & Exponential moving average that forms $R_t^{(k)}$ from allocation reward. \\
\addlinespace[2pt]
$\lambda_{\min}$ & Cost penalty floor & $0.25$ & Early-run lower bound on the cost penalty in $d_t$. \\
\addlinespace[2pt]
$c_{\mathrm{ucb}}$ & Optimism scale & $0.5$ & Frontier exploration bonus scaled by the remaining-budget ratio. \\
\addlinespace[2pt]
$\epsilon_g$ & Meaningful progress threshold & $10^{-4}$ & Stagnation reset and detection of low-yield steps. \\
\addlinespace[2pt]
$\epsilon_{\mathrm{strong}}$ & Substantial improvement threshold & $10^{-3}$ & Guide success, consolidation window, and default refinement mode, with $\epsilon_{\mathrm{strong}}=10\epsilon_g$. \\
\addlinespace[2pt]
$\nu_0$ & Base patience scale & $6$ & Budget-dependent stagnation requirement $\nu_{\mathrm{req}}(\rho_t,K_t)$. \\
\addlinespace[2pt]
$\epsilon_c,\ \epsilon_H$ & Numerical stabilizers & $10^{-8},\ 10^{-8}$ & Cost normalization and the local intensity denominator. \\
\addlinespace[2pt]
$\hat{c}_{\mathrm{guide}}$ & Guide cost estimate & $2\bar{c}$ & Low-yield threshold and guide-cost term of the affordability check. \\
\bottomrule
\end{tabular}
\end{adjustbox}
\end{center}
\end{table}

\subsection{Local Exploration Control}
\label{app:local_exploration_control}

The local controller favors high-scoring parents in exploitation mode, novel parents in exploration mode, and a mixture of the two in balanced mode. Here, novelty measures code dissimilarity relative to candidates in the local archive. Initial sampling uses exploration for the first search step and balanced sampling when another frontier is first selected. For a previously selected frontier, the fixed sampler $\pi_{\mathrm{local}}$ uses the intensity computed from $H_{t-1}^{(k_t)}$ and $\rho_{t-1}$ in Section~\ref{subsec:budget_aware_control}. A consolidation window takes precedence and uses exploitation. Outside initial sampling and consolidation, the probabilities are
\[
\begin{aligned}
\Pr(a_t=\mathrm{exploration})&=I_t^{(k_t)},\\
\Pr(a_t=\mathrm{exploitation})&=0.7(1-I_t^{(k_t)}),\\
\Pr(a_t=\mathrm{balanced})&=0.3(1-I_t^{(k_t)}).
\end{aligned}
\]

The mode changes parent selection and prompt guidance but leaves the context request unchanged. Every mode requests four context programs in addition to the parent. Two are high-scoring local candidates chosen for dissimilarity to the parent, and two are the highest-scoring candidates across frontiers. Retrieval excludes the sampled parent and returns fewer context programs when too few candidates are available. During an active tactic batch, the global best candidate replaces the sampled parent, while the retrieved context is retained.

\subsection{Frontier Allocation Control}
\label{app:frontier_allocation_control}

The allocation state records only global progress after cost calibration, so local-only improvement receives no allocation reward. A frontier with only local improvement can still be selected through the exploration bonus. We update this state only for the selected frontier, leaving the other frontier states unchanged. The allocation rule in Section~\ref{subsec:budget_aware_control} selects the next frontier unless a consolidation window directs search to the successful guide's originating frontier. The shared archive scaffold starts with two frontiers and periodically exchanges high-scoring candidates between them. No new frontier was opened in the reported main runs, so $K_t=2$ throughout.

\subsection{Budgeted Tactic Intervention}
\label{app:budgeted_tactic_intervention}

Budgeted tactic intervention purchases guidance after global stagnation or accumulated low-yield spending, provided that the remaining budget covers the estimated guide cost and a testing reserve. The following rules specify this decision and how guide outcomes affect subsequent search.

\subsubsection{Stagnation and Low-Yield Evidence}
The stagnation counter $\nu_t$ counts consecutive steps with global gain $g_t\le\epsilon_g$ and resets to zero when $g_t>\epsilon_g$. The stagnation indicator compares this count with a budget-adaptive patience requirement:
\[
P_t=[\nu_t\ge\nu_{\mathrm{req}}(\rho_t,K_t)],
\qquad
\nu_{\mathrm{req}}(\rho_t,K_t)
=\max\left(\lceil\nu_0(1-\rho_t)\rceil,2K_t\right),
\]
where $[\cdot]$ equals one when its condition holds and zero otherwise, and $\nu_0$ is the base patience scale. The first term raises the requirement as the budget is spent, so guidance late in the run needs longer stagnation evidence. The lower bound requires at least $2K_t$ consecutive steps without meaningful global improvement, twice the frontier count.

Low-yield spending provides a complementary trigger before the longer stagnation requirement is met. The accumulator $L_t$ sums realized costs $c_t$ over consecutive steps satisfying both $u_t^{(k_t)}\le\epsilon_g$ and $r_t^{(k_t)}\le\epsilon_g$, resetting to zero otherwise. This route requires at least half the budget to remain and at least $K_t$ steps without meaningful global improvement, while the stagnation route remains available later. Using the estimated guide cost $\hat{c}_{\mathrm{guide}}$, the low-yield indicator is
\[
Y_t=[\rho_t\ge\nicefrac{1}{2}]\land[\nu_t\ge K_t]\land[L_t\ge\hat{c}_{\mathrm{guide}}].
\]

\subsubsection{Guide-Cycle Credit}
A guide requests three tactics, which are tested in round-robin order with at most two generation attempts per returned tactic, including retries. When a valid batch is generated, its guide cycle keeps a credit record $\mathcal{G}$ that stores the current global best $y_{\mathrm{base}}^{\mathcal{G}}$, originating frontier $k^{\mathcal{G}}$, and guide mode. While this record is open, accumulated improvement and guide success are measured by
\[
g_t^{\mathcal{G}}
=\max\left(
\frac{y_t-y_{\mathrm{base}}^{\mathcal{G}}}
{\max(|y_t|,|y_{\mathrm{base}}^{\mathcal{G}}|,1)},0
\right),
\qquad
\zeta_t=[g_t^{\mathcal{G}}>\epsilon_{\mathrm{strong}}],
\]
where $\epsilon_{\mathrm{strong}}=10\epsilon_g$. We set $g_t^{\mathcal{G}}=0$ when no credit record is open. The credit record closes at the first success or when the tactics are exhausted without success. Remaining tactics can still be used after success, so $G_t$ may remain one while the post-step credit record $\mathcal{G}_t$ is empty. These later tests do not generate another success event for the resolved record.

\subsubsection{Guide Outcomes}
A successful guide cycle sets $Z_t=1$ and opens a consolidation window of $K_t$ subsequent iterations on its originating frontier. During this window, $\kappa_t$ identifies that frontier and $w_t$ counts the remaining iterations, which use exploitation and suppress new guidance. Success also resets $L_t$ to zero and clears the temporary mode correction and backoff. Remaining tactics can still be tested within this window, but later improvements do not restart it.

If the batch is exhausted without success, new guidance is suppressed on that step and for the next $K_t$ iterations. The backoff counter $\omega_t$ tracks the subsequent pause while candidate search and updates to $\nu_t$ and $L_t$ continue. We choose the next guide mode $m_t$ after these outcome updates. The default is breakthrough while $Z_t=0$ and refinement once $Z_t=1$. A failed refinement cycle sets the temporary correction $o_t$ to request one breakthrough probe. If that probe fails, the correction clears and refinement resumes. When present, the updated correction overrides the default mode.

\subsubsection{Affordability and Intervention Decision}
Guidance requires budget for the estimated guide cost and a testing reserve of $K_t\bar{c}$, the reference cost of $K_t$ ordinary generations. The affordability indicator is therefore
\[
A_t=[B-C_t\ge\hat{c}_{\mathrm{guide}}+K_t\bar{c}].
\]
With guidance enabled, the decision combines affordability with stagnation or low-yield evidence:
\[
\mathcal{I}_t=A_t\land(P_t\lor Y_t),
\]
with $\mathcal{I}_t=0$ during consolidation, while tactics or unresolved guide credit remain, or during the backoff interval defined above. When $\mathcal{I}_t=1$, mode $m_t$ is attached to the next eligible guide call. Appendix~\ref{app:worked_example} traces both trigger paths and both modes in one controller run.

\subsection{Controller Trace}
\label{app:worked_example}

Table~\ref{tab:costada-controller-trace} reports the controller state over iterations 6--20 of the CostAda run plotted in Figure~\ref{fig:glm5-signal-processing-budget-trajectories}. The search budget is $B{=}1$, and the reference cost is $\bar{c}=0.01$. The window shows the scheduling of three guide calls and the global improvements that follow. Within this window, uneven action costs and local-only gains produce distinct utility and allocation signals. Local intensity and guide timing reflect both recent progress and the remaining budget.

\begin{table}[!htbp]
\caption{\textbf{Controller trace over iterations 6--20 on Signal Processing under GLM-5.} Parentheses give local sampling intensity, and $\rho_t$ is the post-step remaining-budget ratio. Utility follows Eq.~\eqref{eq:utility}. Step cost includes guide/tactic calls issued in that iteration. Blank Event cells indicate no controller event during the corresponding search iteration.}
\label{tab:costada-controller-trace}
\begin{center}
\begin{adjustbox}{max width=\linewidth}
\begin{tabular}{c c l c c c c l}
\toprule
$t$ & $k_t$ & Local action & $c_t$ & $\rho_t$ & $u_t$ & $r_t$ & Event \\
\midrule
6 & 1 & exploration (0.46) & 0.0073 & 0.92 & 0.000 & 0.000 & stagnation met, guide scheduled (breakthrough) \\
7 & 2 & balanced (0.46) & 0.0179 & 0.90 & 0.056 & 0.056 & guide charged \$0.0084, consolidation opens \\
8 & 2 & exploitation (0.15) & 0.0309 & 0.87 & 0.000 & 0.000 & consolidation \\
9 & 2 & exploitation (0.15) & 0.0266 & 0.84 & 0.000 & 0.000 & consolidation \\
10 & 1 & exploration (0.44) & 0.0085 & 0.83 & 0.050 & 0.000 & local gain, no global gain \\
11 & 1 & exploration (0.42) & 0.0160 & 0.82 & 0.000 & 0.000 & \\
12 & 1 & exploration (0.42) & 0.0151 & 0.80 & 0.000 & 0.000 & stagnation met, guide scheduled (refinement) \\
13 & 2 & exploration (0.41) & 0.0313 & 0.77 & 0.000 & 0.000 & guide charged \$0.0087 \\
14 & 1 & balanced (0.40) & 0.0149 & 0.76 & 0.000 & 0.000 & \\
15 & 2 & balanced (0.40) & 0.0255 & 0.73 & 0.052 & 0.052 & consolidation opens \\
16 & 2 & exploitation (0.15) & 0.0196 & 0.71 & 0.000 & 0.000 & consolidation \\
17 & 2 & exploitation (0.15) & 0.0081 & 0.70 & 0.000 & 0.000 & low-yield met, guide scheduled (refinement) \\
18 & 1 & exploration (0.38) & 0.0361 & 0.67 & 0.000 & 0.000 & guide charged \$0.0116 \\
19 & 1 & exploration (0.37) & 0.0146 & 0.65 & 0.000 & 0.000 & \\
20 & 1 & exploration (0.37) & 0.0144 & 0.64 & 0.062 & 0.062 & consolidation opens \\
\bottomrule
\end{tabular}
\end{adjustbox}
\end{center}
\end{table}

First, realized action costs vary substantially within the same run. Step costs inside the displayed window range from \$0.0073 to \$0.0361, making the costliest step about five times the cheapest. Across the full run, the seven guide/tactic calls cost between \$0.0074 and \$0.0179 each. This variation also appears in the middle panel of Figure~\ref{fig:glm5-signal-processing-budget-trajectories}. Through the cost-calibration factor $d_t$, more expensive actions receive less credit for a fixed progress mixture and remaining budget.

Second, local refinement and global improvement produce different evidence for subsequent control. At iteration 10, a normalized local gain of $0.0692$ improves the frontier archive without improving the global best. Equation~\eqref{eq:utility} assigns a positive cost-calibrated utility of $0.050$. The allocation reward is zero because frontier allocation credits only global gain. The positive utility updates $H_t^{(k_t)}$ and thereby affects local sampling intensity, without adding global-gain reward to the allocation state. By contrast, local and global gains coincide at iterations 7, 15, and 20. Utility and allocation reward are equal at $0.056$, $0.052$, and $0.062$, respectively.

Local control combines gradual narrowing as the budget shrinks with temporary exploitation during consolidation. Intensity falls from $0.46$ at $\rho_t=0.92$ to $0.37$ by the end of the displayed window. Later in the run, it reaches $0.26$ at $\rho_t=0.32$ and $0.15$ at exhaustion. At iterations 8--9 and 16--17, consolidation forces exploitation with intensity $0.15$, producing sharper changes within the gradual decline. Sampling returns to exploration when each consolidation window closes, linking these temporary overrides to successful guide cycles.

Intervention responds to both stagnation and low-yield spending, while previous guide outcomes determine whether its default mode seeks breakthrough or refinement. At iteration 6, the global stagnation counter reaches its requirement of four. CostAda schedules a breakthrough guide because no earlier guide has produced substantial global improvement. The guide is charged \$0.0084 at iteration 7 and is followed by a normalized global gain of $0.0700$, which opens a consolidation window. After this success, the guides scheduled at iterations 12 and 17 use refinement mode. Their triggers differ, with stagnation prompting the guide at iteration 12 and low-yield spending prompting it at iteration 17, when $\rho_t=0.70$. By iteration 17, the run has accumulated \$0.0277 of low-yield spending, above the estimated guide cost of \$0.02 ($2\bar{c}$). The stagnation count of two meets the low-yield requirement of $K_t=2$ steps but remains below the stagnation requirement of four. Accumulated spending therefore triggers the earlier guide schedule.

Near budget exhaustion, CostAda increases both the required stagnation evidence and the cost penalty on progress. The required stagnation count remains four throughout the displayed window. It rises to five when $\rho_t$ reaches $0.32$ and to six by $\rho_t=0.16$ because patience scales with $1-\rho_t$. The cost-calibration factor $d_t$ rises from about $1.14$ early in the run to $1.72$ at $\rho_t=0.32$, $2.01$ at $\rho_t=0.16$, and $2.48$ at $\rho_t=0.02$. These values reflect both realized step costs and the increasing weight $\lambda_t^c$. For a fixed progress mixture and realized cost, a larger $\lambda_t^c$ reduces utility as the remaining budget shrinks. The run ends with final spending of \$1.0090, exceeding nominal $B{=}1$ by $0.90\%$.

\section{Benchmark Details}
\label{app:benchmark_details}

We follow the benchmark definitions and evaluation setup used by SkyDiscover, AdaEvolve, and EvoX \citep{liu2026skydiscover,cemri2026adaevolve,liu2026evox}. The underlying tasks and reference values draw on AlphaEvolve, packing records, Heilbronn's problem, generalized Sidon sets, OpenEvolve, and standard signal-processing references \citep{novikov2025alphaevolve,friedman2025packing,roth1951heilbronn,vinuesa2010generalized,openevolve,shenoi2005introduction}. The controller uses the higher-is-better evaluator score $F(p)$ to compute local and global gains in Section~\ref{sec:method}. Table~\ref{tab:math8-benchmark-details} lists each benchmark's native objective and optimization direction for reporting results. Appendix~\ref{app:discovered_programs} shows final CostAda programs for Circle Packing and MinMaxDist $(n{=}16,d{=}2)$.

\begin{table}[!htbp]
\caption{
\textbf{Benchmark objectives and optimization directions.}
Rows state the benchmark objective and whether higher or lower values are preferred.
}
\label{tab:math8-benchmark-details}
\begin{center}
\begin{adjustbox}{max width=\linewidth}
\begin{tabular}{p{3.5cm} p{1.4cm} p{10.5cm}}
\toprule
\textbf{Benchmark} & \textbf{Direction} & \textbf{Objective} \\
\midrule
Circle Packing & $\uparrow$ & Pack $26$ disjoint circles in a unit square and maximize the sum of radii. \\
Circle Packing Rect & $\uparrow$ & Pack $21$ disjoint circles in a rectangle of perimeter $4$ and maximize the sum of radii over all feasible circle configurations. \\
Heilbronn Triangle & $\uparrow$ & Place $11$ points in a unit area triangle and maximize the minimum area over all triangles formed by triples of points. \\
Heilbronn Convex & $\uparrow$ & Place $13$ points in a unit area convex region and maximize the minimum triangle area induced by any three points. \\
MinMaxDist $(n{=}16,d{=}2)$ & $\downarrow$ & Place $16$ points in two dimensions and minimize the ratio between maximum and minimum squared pairwise distances \citep{novikov2025alphaevolve}. \\
MinMaxDist $(n{=}14,d{=}3)$ & $\downarrow$ & Place $14$ points in three dimensions and minimize the ratio between maximum and minimum squared pairwise distances. \\
Third Autocorrelation & $\downarrow$ & Construct witness functions that improve the upper bound on the third autocorrelation constant $C_3$ in additive combinatorics. \\
Signal Processing & $\uparrow$ & Synthesize a filtering program for noisy nonstationary time series, balancing fidelity, smoothness, lag, and false trend changes. \\
\bottomrule
\end{tabular}
\end{adjustbox}
\end{center}
\end{table}

For Signal Processing, candidates are ranked by \texttt{combined\_score}, which is identical to \texttt{overall\_score}. The evaluator combines the original composite score, smoothness score, correlation/accuracy score, noise-reduction score, and success rate with weights $0.4$, $0.2$, $0.2$, $0.1$, and $0.1$, respectively. The score is set to zero when the correlation/accuracy score is below $0.1$. We use this score for all Signal Processing cutoff and normalized-\textsc{AUC} results.

\section{Experimental Accounting Details}
\label{app:experimental_accounting}

\subsection{Search-Side Cost}
\label{app:search_side_cost}

If iteration $t$ issues a set of search-side LLM calls $\mathcal{J}_t$, its realized cost is
\[
c_t
=
\sum_{j\in\mathcal{J}_t}
\left(
p_{\mathrm{in}}^{(\mu_j)}T_{\mathrm{in}}^{(j)}
+
p_{\mathrm{out}}^{(\mu_j)}T_{\mathrm{out}}^{(j)}
\right),
\]
where $\mu_j$ is the model used by call $j$, $T_{\mathrm{in}}^{(j)}$ and $T_{\mathrm{out}}^{(j)}$ are provider-reported input and output token counts, and $p_{\mathrm{in}}^{(\mu_j)}$, $p_{\mathrm{out}}^{(\mu_j)}$ are the corresponding fixed per-token prices. We use the same OpenRouter prices for all methods within each backbone. GLM-5 input/output prices are \$0.60/\$1.92 per million tokens \citep{openrouter2026glm5pricing}, and GPT-5.4 prices are \$2.50/\$15.00 per million tokens \citep{openrouter2026gpt54pricing}. We apply these fixed rates without cache discounts.

Search-side calls include ordinary candidate generation, recovery from failed generations, and guide/tactic generation. The scalar controller updates require no additional LLM calls. Their CPU time and deterministic evaluator execution are excluded. All reported budgets use this ledger.

The normalized step cost $\tilde{c}_t$ uses reference generation costs of $\bar{c}{=}0.01$ for GLM-5 and $\bar{c}{=}0.1$ for GPT-5.4. We measured these values before the main comparisons as the average realized search-side cost of an ordinary generation call. The reference cost is fixed per backbone and shared across benchmarks. It scales the controller's cost credit without changing the monetary evaluation ledger.

\subsection{Shared Experimental Scaffold}
\label{app:shared_experimental_scaffold}

We instantiate all methods in the same discovery framework. Candidate-generation infrastructure, parsing, deterministic evaluator execution, and archive insertion are shared across methods. For the main comparisons, generator and guide model choices are fixed across all eight benchmarks under each backbone. Every method uses the same fixed prices.

\section{Budget-Cutoff Results}
\label{app:additional_budget_cutoffs}

The cutoff results below support the budget-path analysis in Section~\ref{subsec:budget_cutoffs}. Tables~\ref{tab:glm5-budget-cutoffs} and~\ref{tab:gpt54-budget-cutoffs} report the results at the budget cutoffs $0.25B$, $0.5B$, $0.75B$, and $B$ under GLM-5 and GPT-5.4, respectively. We evaluate all three methods at the same four cutoffs. We omit the zero-cost initialization because it precedes candidate generation and may lack a benchmark-objective value.

\begin{table}[!htb]
\caption{\textbf{Benchmark objectives across GLM-5 budget cutoffs.} With search budget $B{=}1$, the columns correspond to costs $0.25$, $0.5$, $0.75$, and $1.0$ in ledger dollars. Entries report \textbf{Mean $\pm$ Std} in the units of Table~\ref{tab:costada-math8-final-quality}. Arrows indicate the preferred direction. Best entries per benchmark and cutoff are bolded, including ties at the reported precision.}
\label{tab:glm5-budget-cutoffs}
\begin{center}
\begin{adjustbox}{max width=\linewidth}
\begin{tabular}{l l c c c c}
\toprule
\textbf{Benchmark} & \textbf{Method} & \textbf{\boldmath$0.25B$} & \textbf{\boldmath$0.5B$} & \textbf{\boldmath$0.75B$} & \textbf{\boldmath$B$} \\
\midrule
\multirow{3}{*}{Circle Packing $\uparrow$}
& AdaEvolve & \meanstd{1.8225 $\pm$ 0.2223} & \meanstd{2.4690 $\pm$ 0.1562} & \meanstd{2.5617 $\pm$ 0.0773} & \meanstd{2.5617 $\pm$ 0.0773} \\
& EvoX & \meanstd{2.1815 $\pm$ 0.0801} & \meanstd{2.5128 $\pm$ 0.1269} & \meanstd{2.5717 $\pm$ 0.0752} & \meanstd{2.5717 $\pm$ 0.0752} \\
& CostAda & \textbf{\meanstd{2.5535 $\pm$ 0.0585}} & \textbf{\meanstd{2.6016 $\pm$ 0.0335}} & \textbf{\meanstd{2.6252 $\pm$ 0.0094}} & \textbf{\meanstd{2.6306 $\pm$ 0.0054}} \\
\midrule
\multirow{3}{*}{Circle Packing Rect $\uparrow$}
& AdaEvolve & \meanstd{2.1182 $\pm$ 0.1146} & \meanstd{2.2043 $\pm$ 0.1578} & \meanstd{2.2089 $\pm$ 0.1635} & \meanstd{2.2089 $\pm$ 0.1635} \\
& EvoX & \meanstd{2.1985 $\pm$ 0.2569} & \meanstd{2.1989 $\pm$ 0.2563} & \meanstd{2.1989 $\pm$ 0.2563} & \meanstd{2.1989 $\pm$ 0.2563} \\
& CostAda & \textbf{\meanstd{2.3478 $\pm$ 0.0081}} & \textbf{\meanstd{2.3478 $\pm$ 0.0081}} & \textbf{\meanstd{2.3478 $\pm$ 0.0081}} & \textbf{\meanstd{2.3525 $\pm$ 0.0081}} \\
\midrule
\multirow{3}{*}{Heilbronn Convex $\uparrow$}
& AdaEvolve & \textbf{\meanstd{0.0215 $\pm$ 0.0030}} & \meanstd{0.0215 $\pm$ 0.0030} & \meanstd{0.0229 $\pm$ 0.0015} & \meanstd{0.0229 $\pm$ 0.0015} \\
& EvoX & \meanstd{0.0203 $\pm$ 0.0036} & \meanstd{0.0222 $\pm$ 0.0024} & \meanstd{0.0234 $\pm$ 0.0012} & \meanstd{0.0235 $\pm$ 0.0012} \\
& CostAda & \textbf{\meanstd{0.0215 $\pm$ 0.0003}} & \textbf{\meanstd{0.0250 $\pm$ 0.0030}} & \textbf{\meanstd{0.0253 $\pm$ 0.0033}} & \textbf{\meanstd{0.0263 $\pm$ 0.0018}} \\
\midrule
\multirow{3}{*}{Heilbronn Triangle $\uparrow$}
& AdaEvolve & \meanstd{0.0212 $\pm$ 0.0075} & \meanstd{0.0283 $\pm$ 0.0020} & \meanstd{0.0283 $\pm$ 0.0020} & \meanstd{0.0303 $\pm$ 0.0011} \\
& EvoX & \meanstd{0.0149 $\pm$ 0.0027} & \meanstd{0.0200 $\pm$ 0.0038} & \meanstd{0.0200 $\pm$ 0.0038} & \meanstd{0.0220 $\pm$ 0.0050} \\
& CostAda & \textbf{\meanstd{0.0296 $\pm$ 0.0017}} & \textbf{\meanstd{0.0296 $\pm$ 0.0017}} & \textbf{\meanstd{0.0296 $\pm$ 0.0017}} & \textbf{\meanstd{0.0314 $\pm$ 0.0003}} \\
\midrule
\multirow{3}{*}{MinMaxDist $(n{=}16,d{=}2)$ $\downarrow$}
& AdaEvolve & \meanstd{14.59 $\pm$ 1.03} & \meanstd{14.54 $\pm$ 1.02} & \meanstd{14.06 $\pm$ 1.30} & \meanstd{14.06 $\pm$ 1.30} \\
& EvoX & \meanstd{15.01 $\pm$ 1.69} & \meanstd{14.37 $\pm$ 2.11} & \meanstd{14.20 $\pm$ 1.90} & \meanstd{14.17 $\pm$ 1.92} \\
& CostAda & \textbf{\meanstd{13.62 $\pm$ 1.02}} & \textbf{\meanstd{13.57 $\pm$ 1.07}} & \textbf{\meanstd{13.15 $\pm$ 0.36}} & \textbf{\meanstd{13.11 $\pm$ 0.29}} \\
\midrule
\multirow{3}{*}{MinMaxDist $(n{=}14,d{=}3)$ $\downarrow$}
& AdaEvolve & \meanstd{4.57 $\pm$ 0.08} & \meanstd{4.48 $\pm$ 0.13} & \meanstd{4.38 $\pm$ 0.08} & \meanstd{4.38 $\pm$ 0.08} \\
& EvoX & \meanstd{4.59 $\pm$ 0.37} & \meanstd{4.34 $\pm$ 0.27} & \meanstd{4.22 $\pm$ 0.06} & \meanstd{4.20 $\pm$ 0.03} \\
& CostAda & \textbf{\meanstd{4.31 $\pm$ 0.24}} & \textbf{\meanstd{4.19 $\pm$ 0.04}} & \textbf{\meanstd{4.18 $\pm$ 0.02}} & \textbf{\meanstd{4.17 $\pm$ 0.00}} \\
\midrule
\multirow{3}{*}{Third Autocorrelation $\downarrow$}
& AdaEvolve & \textbf{\meanstd{1.4711 $\pm$ 0.0040}} & \textbf{\meanstd{1.4694 $\pm$ 0.0017}} & \meanstd{1.4694 $\pm$ 0.0017} & \meanstd{1.4685 $\pm$ 0.0014} \\
& EvoX & \meanstd{1.5760 $\pm$ 0.1729} & \meanstd{1.5719 $\pm$ 0.1658} & \meanstd{1.5481 $\pm$ 0.1308} & \meanstd{1.5481 $\pm$ 0.1308} \\
& CostAda & \meanstd{1.4858 $\pm$ 0.0256} & \meanstd{1.4746 $\pm$ 0.0064} & \textbf{\meanstd{1.4679 $\pm$ 0.0054}} & \textbf{\meanstd{1.4658 $\pm$ 0.0037}} \\
\midrule
\multirow{3}{*}{Signal Processing $\uparrow$}
& AdaEvolve & \meanstd{0.5692 $\pm$ 0.0301} & \meanstd{0.6126 $\pm$ 0.0161} & \meanstd{0.6173 $\pm$ 0.0158} & \meanstd{0.6393 $\pm$ 0.0240} \\
& EvoX & \meanstd{0.5122 $\pm$ 0.0177} & \meanstd{0.6361 $\pm$ 0.0655} & \meanstd{0.6429 $\pm$ 0.0544} & \meanstd{0.6502 $\pm$ 0.0430} \\
& CostAda & \textbf{\meanstd{0.6377 $\pm$ 0.0805}} & \textbf{\meanstd{0.6924 $\pm$ 0.0536}} & \textbf{\meanstd{0.7069 $\pm$ 0.0318}} & \textbf{\meanstd{0.7222 $\pm$ 0.0127}} \\
\bottomrule
\end{tabular}
\end{adjustbox}
\end{center}
\end{table}

\begin{table}[!htbp]
\caption{\textbf{Benchmark objectives across GPT-5.4 budget cutoffs.} With search budget $B{=}5$, the columns correspond to costs $1.25$, $2.5$, $3.75$, and $5.0$ in ledger dollars. Entries report \textbf{Mean $\pm$ Std} in the units of Table~\ref{tab:costada-math8-final-quality}. Arrows indicate the preferred direction. Best entries per benchmark and cutoff are bolded, including ties at the reported precision.}
\label{tab:gpt54-budget-cutoffs}
\begin{center}
\begin{adjustbox}{max width=\linewidth}
\begin{tabular}{l l c c c c}
\toprule
\textbf{Benchmark} & \textbf{Method} & \textbf{\boldmath$0.25B$} & \textbf{\boldmath$0.5B$} & \textbf{\boldmath$0.75B$} & \textbf{\boldmath$B$} \\
\midrule
\multirow{3}{*}{Circle Packing $\uparrow$}
& AdaEvolve & \meanstd{2.5632 $\pm$ 0.0784} & \meanstd{2.5827 $\pm$ 0.0510} & \meanstd{2.5865 $\pm$ 0.0535} & \meanstd{2.6196 $\pm$ 0.0030} \\
& EvoX & \meanstd{2.5225 $\pm$ 0.0328} & \meanstd{2.5441 $\pm$ 0.0609} & \meanstd{2.5818 $\pm$ 0.0476} & \meanstd{2.5851 $\pm$ 0.0510} \\
& CostAda & \textbf{\meanstd{2.6143 $\pm$ 0.0140}} & \textbf{\meanstd{2.6211 $\pm$ 0.0033}} & \textbf{\meanstd{2.6218 $\pm$ 0.0044}} & \textbf{\meanstd{2.6248 $\pm$ 0.0046}} \\
\midrule
\multirow{3}{*}{Circle Packing Rect $\uparrow$}
& AdaEvolve & \meanstd{2.2826 $\pm$ 0.0410} & \meanstd{2.2910 $\pm$ 0.0507} & \meanstd{2.3125 $\pm$ 0.0370} & \meanstd{2.3148 $\pm$ 0.0357} \\
& EvoX & \textbf{\meanstd{2.3221 $\pm$ 0.0428}} & \textbf{\meanstd{2.3474 $\pm$ 0.0011}} & \meanstd{2.3474 $\pm$ 0.0011} & \meanstd{2.3474 $\pm$ 0.0011} \\
& CostAda & \meanstd{2.2689 $\pm$ 0.0313} & \meanstd{2.3111 $\pm$ 0.0462} & \textbf{\meanstd{2.3491 $\pm$ 0.0128}} & \textbf{\meanstd{2.3577 $\pm$ 0.0050}} \\
\midrule
\multirow{3}{*}{Heilbronn Convex $\uparrow$}
& AdaEvolve & \meanstd{0.0187 $\pm$ 0.0007} & \meanstd{0.0187 $\pm$ 0.0007} & \meanstd{0.0187 $\pm$ 0.0007} & \meanstd{0.0197 $\pm$ 0.0020} \\
& EvoX & \meanstd{0.0166 $\pm$ 0.0018} & \meanstd{0.0197 $\pm$ 0.0026} & \meanstd{0.0212 $\pm$ 0.0021} & \meanstd{0.0223 $\pm$ 0.0031} \\
& CostAda & \textbf{\meanstd{0.0234 $\pm$ 0.0010}} & \textbf{\meanstd{0.0234 $\pm$ 0.0010}} & \textbf{\meanstd{0.0244 $\pm$ 0.0008}} & \textbf{\meanstd{0.0244 $\pm$ 0.0008}} \\
\midrule
\multirow{3}{*}{Heilbronn Triangle $\uparrow$}
& AdaEvolve & \meanstd{0.0280 $\pm$ 0.0003} & \meanstd{0.0290 $\pm$ 0.0016} & \meanstd{0.0299 $\pm$ 0.0014} & \meanstd{0.0307 $\pm$ 0.0008} \\
& EvoX & \meanstd{0.0240 $\pm$ 0.0022} & \meanstd{0.0254 $\pm$ 0.0019} & \meanstd{0.0260 $\pm$ 0.0030} & \meanstd{0.0263 $\pm$ 0.0035} \\
& CostAda & \textbf{\meanstd{0.0288 $\pm$ 0.0039}} & \textbf{\meanstd{0.0317 $\pm$ 0.0014}} & \textbf{\meanstd{0.0329 $\pm$ 0.0012}} & \textbf{\meanstd{0.0330 $\pm$ 0.0009}} \\
\midrule
\multirow{3}{*}{MinMaxDist $(n{=}16,d{=}2)$ $\downarrow$}
& AdaEvolve & \meanstd{13.49 $\pm$ 0.85} & \meanstd{13.37 $\pm$ 0.65} & \meanstd{13.14 $\pm$ 0.24} & \meanstd{13.00 $\pm$ 0.00} \\
& EvoX & \textbf{\meanstd{13.00 $\pm$ 0.00}} & \meanstd{13.00 $\pm$ 0.00} & \meanstd{13.00 $\pm$ 0.00} & \meanstd{13.00 $\pm$ 0.00} \\
& CostAda & \meanstd{13.16 $\pm$ 0.20} & \textbf{\meanstd{12.98 $\pm$ 0.09}} & \textbf{\meanstd{12.93 $\pm$ 0.06}} & \textbf{\meanstd{12.93 $\pm$ 0.06}} \\
\midrule
\multirow{3}{*}{MinMaxDist $(n{=}14,d{=}3)$ $\downarrow$}
& AdaEvolve & \meanstd{4.64 $\pm$ 0.15} & \meanstd{4.61 $\pm$ 0.14} & \meanstd{4.61 $\pm$ 0.14} & \meanstd{4.55 $\pm$ 0.15} \\
& EvoX & \meanstd{4.64 $\pm$ 0.16} & \meanstd{4.46 $\pm$ 0.01} & \meanstd{4.44 $\pm$ 0.02} & \meanstd{4.44 $\pm$ 0.02} \\
& CostAda & \textbf{\meanstd{4.55 $\pm$ 0.15}} & \textbf{\meanstd{4.42 $\pm$ 0.08}} & \textbf{\meanstd{4.42 $\pm$ 0.08}} & \textbf{\meanstd{4.42 $\pm$ 0.08}} \\
\midrule
\multirow{3}{*}{Third Autocorrelation $\downarrow$}
& AdaEvolve & \meanstd{1.5433 $\pm$ 0.0278} & \meanstd{1.5157 $\pm$ 0.0079} & \meanstd{1.5157 $\pm$ 0.0079} & \meanstd{1.5081 $\pm$ 0.0069} \\
& EvoX & \meanstd{1.4899 $\pm$ 0.0028} & \meanstd{1.4853 $\pm$ 0.0043} & \meanstd{1.4838 $\pm$ 0.0029} & \meanstd{1.4824 $\pm$ 0.0042} \\
& CostAda & \textbf{\meanstd{1.4748 $\pm$ 0.0056}} & \textbf{\meanstd{1.4738 $\pm$ 0.0048}} & \textbf{\meanstd{1.4717 $\pm$ 0.0074}} & \textbf{\meanstd{1.4717 $\pm$ 0.0074}} \\
\midrule
\multirow{3}{*}{Signal Processing $\uparrow$}
& AdaEvolve & \meanstd{0.6617 $\pm$ 0.0040} & \meanstd{0.6739 $\pm$ 0.0145} & \meanstd{0.7043 $\pm$ 0.0024} & \meanstd{0.7182 $\pm$ 0.0137} \\
& EvoX & \meanstd{0.5890 $\pm$ 0.0447} & \meanstd{0.6436 $\pm$ 0.0509} & \meanstd{0.6644 $\pm$ 0.0299} & \meanstd{0.6767 $\pm$ 0.0179} \\
& CostAda & \textbf{\meanstd{0.7069 $\pm$ 0.0402}} & \textbf{\meanstd{0.7289 $\pm$ 0.0262}} & \textbf{\meanstd{0.7289 $\pm$ 0.0262}} & \textbf{\meanstd{0.7672 $\pm$ 0.0292}} \\
\bottomrule
\end{tabular}
\end{adjustbox}
\end{center}
\end{table}

In Tables~\ref{tab:glm5-budget-cutoffs} and~\ref{tab:gpt54-budget-cutoffs}, CostAda reaches the strongest baseline's full-budget quality with at most half the budget on six of the eight benchmarks under GLM-5 and seven under \mbox{GPT-5.4}. CostAda also leads or ties 30 of the 32 benchmark-objective cells under GLM-5. Under \mbox{GPT-5.4}, it leads 29 of the 32 cells. From $0.75B$ onward, it leads on all eight benchmarks under both backbones. These per-benchmark results agree with the early and persistent lead in Figure~\ref{fig:budget-path-summary}.

\section{Small-Budget Evaluation}
\label{app:small_budget_rerun_vs_cutoff}

To validate the early-budget comparisons under GLM-5, we additionally evaluate CostAda in independent runs with a smaller search budget (Table~\ref{tab:glm5-small-budget-rerun-vs-cutoff}). Within each benchmark block, the first three rows report cutoff values from $B{=}1$ trajectories at the corresponding absolute costs. CostAda$_{B=0.5}$ instead reports three independent runs configured with search budget $B{=}0.5$ and evaluated at the same absolute costs. These runs can adapt their search to the smaller horizon, whereas a cutoff only truncates a trajectory planned for $B{=}1$.

\begin{table}[!htbp]
\caption{\textbf{Benchmark objectives for independent $B{=}0.5$ runs and $B{=}1$ trajectories under \mbox{GLM-5}.}
The first three rows of each benchmark block come from $B{=}1$ trajectories. CostAda$_{B=0.5}$ denotes runs configured with the smaller search budget and evaluated at the same costs, \$0.25 and \$0.5. Entries report \textbf{Mean $\pm$ Std} in the units of Table~\ref{tab:costada-math8-final-quality}. Arrows indicate the preferred direction. Best entries per benchmark and cutoff are bolded, including ties at the reported precision.}
\label{tab:glm5-small-budget-rerun-vs-cutoff}
\begin{center}
\small
\setlength{\tabcolsep}{4pt}
\renewcommand{\arraystretch}{0.92}
\begin{adjustbox}{max width=\linewidth}
\begin{tabular}{l l c c}
\toprule
\textbf{Benchmark} & \textbf{Method} & \textbf{\boldmath$0.25B$} & \textbf{\boldmath$0.5B$} \\
\midrule
\multirow{4}{*}{Circle Packing $\uparrow$}
& AdaEvolve & \meanstd{1.8225 $\pm$ 0.2223} & \meanstd{2.4690 $\pm$ 0.1562} \\
& EvoX & \meanstd{2.1815 $\pm$ 0.0801} & \meanstd{2.5128 $\pm$ 0.1269} \\
& CostAda & \meanstd{2.5535 $\pm$ 0.0585} & \meanstd{2.6016 $\pm$ 0.0335} \\
& CostAda$_{B=0.5}$ & \textbf{\meanstd{2.6146 $\pm$ 0.0176}} & \textbf{\meanstd{2.6233 $\pm$ 0.0044}} \\
\midrule
\multirow{4}{*}{Circle Packing Rect $\uparrow$}
& AdaEvolve & \meanstd{2.1182 $\pm$ 0.1146} & \meanstd{2.2043 $\pm$ 0.1578} \\
& EvoX & \meanstd{2.1985 $\pm$ 0.2569} & \meanstd{2.1989 $\pm$ 0.2563} \\
& CostAda & \textbf{\meanstd{2.3478 $\pm$ 0.0081}} & \textbf{\meanstd{2.3478 $\pm$ 0.0081}} \\
& CostAda$_{B=0.5}$ & \meanstd{2.3404 $\pm$ 0.0008} & \meanstd{2.3404 $\pm$ 0.0008} \\
\midrule
\multirow{4}{*}{Heilbronn Convex $\uparrow$}
& AdaEvolve & \meanstd{0.0215 $\pm$ 0.0030} & \meanstd{0.0215 $\pm$ 0.0030} \\
& EvoX & \meanstd{0.0203 $\pm$ 0.0036} & \meanstd{0.0222 $\pm$ 0.0024} \\
& CostAda & \meanstd{0.0215 $\pm$ 0.0003} & \meanstd{0.0250 $\pm$ 0.0030} \\
& CostAda$_{B=0.5}$ & \textbf{\meanstd{0.0253 $\pm$ 0.0011}} & \textbf{\meanstd{0.0253 $\pm$ 0.0011}} \\
\midrule
\multirow{4}{*}{Heilbronn Triangle $\uparrow$}
& AdaEvolve & \meanstd{0.0212 $\pm$ 0.0075} & \meanstd{0.0283 $\pm$ 0.0020} \\
& EvoX & \meanstd{0.0149 $\pm$ 0.0027} & \meanstd{0.0200 $\pm$ 0.0038} \\
& CostAda & \meanstd{0.0296 $\pm$ 0.0017} & \meanstd{0.0296 $\pm$ 0.0017} \\
& CostAda$_{B=0.5}$ & \textbf{\meanstd{0.0302 $\pm$ 0.0036}} & \textbf{\meanstd{0.0310 $\pm$ 0.0023}} \\
\midrule
\multirow{4}{*}{MinMaxDist $(n{=}16,d{=}2)$ $\downarrow$}
& AdaEvolve & \meanstd{14.59 $\pm$ 1.03} & \meanstd{14.54 $\pm$ 1.02} \\
& EvoX & \meanstd{15.01 $\pm$ 1.69} & \meanstd{14.37 $\pm$ 2.11} \\
& CostAda & \meanstd{13.62 $\pm$ 1.02} & \meanstd{13.57 $\pm$ 1.07} \\
& CostAda$_{B=0.5}$ & \textbf{\meanstd{13.41 $\pm$ 0.46}} & \textbf{\meanstd{13.13 $\pm$ 0.21}} \\
\midrule
\multirow{4}{*}{MinMaxDist $(n{=}14,d{=}3)$ $\downarrow$}
& AdaEvolve & \meanstd{4.57 $\pm$ 0.08} & \meanstd{4.48 $\pm$ 0.13} \\
& EvoX & \meanstd{4.59 $\pm$ 0.37} & \meanstd{4.34 $\pm$ 0.27} \\
& CostAda & \meanstd{4.31 $\pm$ 0.24} & \meanstd{4.19 $\pm$ 0.04} \\
& CostAda$_{B=0.5}$ & \textbf{\meanstd{4.23 $\pm$ 0.09}} & \textbf{\meanstd{4.17 $\pm$ 0.00}} \\
\midrule
\multirow{4}{*}{Third Autocorrelation $\downarrow$}
& AdaEvolve & \meanstd{1.4711 $\pm$ 0.0040} & \meanstd{1.4694 $\pm$ 0.0017} \\
& EvoX & \meanstd{1.5760 $\pm$ 0.1729} & \meanstd{1.5719 $\pm$ 0.1658} \\
& CostAda & \meanstd{1.4858 $\pm$ 0.0256} & \meanstd{1.4746 $\pm$ 0.0064} \\
& CostAda$_{B=0.5}$ & \textbf{\meanstd{1.4685 $\pm$ 0.0062}} & \textbf{\meanstd{1.4685 $\pm$ 0.0062}} \\
\midrule
\multirow{4}{*}{Signal Processing $\uparrow$}
& AdaEvolve & \meanstd{0.5692 $\pm$ 0.0301} & \meanstd{0.6126 $\pm$ 0.0161} \\
& EvoX & \meanstd{0.5122 $\pm$ 0.0177} & \meanstd{0.6361 $\pm$ 0.0655} \\
& CostAda & \meanstd{0.6377 $\pm$ 0.0805} & \meanstd{0.6924 $\pm$ 0.0536} \\
& CostAda$_{B=0.5}$ & \textbf{\meanstd{0.6690 $\pm$ 0.0095}} & \textbf{\meanstd{0.7120 $\pm$ 0.0169}} \\
\bottomrule
\end{tabular}
\end{adjustbox}
\end{center}
\end{table}

CostAda's early-budget advantage persists when the smaller budget is configured before search. In Table~\ref{tab:glm5-small-budget-rerun-vs-cutoff}, the independently budgeted CostAda$_{B=0.5}$ runs outperform the strongest baseline result in all 16 benchmark--cutoff cells. This comparison is made at the same absolute costs, with baseline values read from $B{=}1$ trajectories. The CostAda$_{B=0.5}$ runs also match or improve on the corresponding CostAda cutoff values from $B{=}1$ trajectories in 14 cells. These independent runs therefore support the budget-cutoff comparisons under GLM-5.

\section{Ablation Study}
\label{app:component_ablations}

We detail the ablation study of Section~\ref{subsec:component_ablations} on Circle Packing, Heilbronn Triangle, and Signal Processing under both backbones, covering packing, geometric point configuration, and executable program synthesis. The study compares CostAda with three variants. The variant w/o cost calibration sets the cost-calibration factor $d_t$ to one while retaining the remaining-budget and intervention rules. The variant w/o budget conditioning fixes $\lambda_t=\lambda_t^c=0.5$ and removes $\rho_t$ from local exploration intensity and the exploration bonus, while leaving intervention gating unchanged. The variant w/o intervention gating schedules guidance from stagnation evidence without the $A_t$ and $Y_t$ gates, while retaining guide backoff, the consolidation window, and the single-active-batch constraint.

For each run, \textsc{BestScore@}$0.5B$ and \textsc{BestScore@}$B$ report the benchmark objective of the best-so-far candidate at the corresponding cutoff, with candidates ranked by normalized score. Normalized \textsc{AUC} is the average stepwise best-so-far normalized score over $[0,B]$, obtained by integrating over realized cost and dividing by $B$. Each action is charged before its score takes effect.

\begin{table}[!htbp]
\caption{
\textbf{Ablation study of CostAda under GLM-5.}
Entries report \textbf{Mean $\pm$ Std}. Benchmark objectives are used for the two cutoff columns, and higher normalized \textsc{AUC} is better. Arrows indicate the preferred direction. Best entries per benchmark and metric among CostAda and its variants are bolded, including ties at the reported precision.
}
\label{tab:glm5-component-ablations}
\begin{center}
\begin{adjustbox}{max width=\linewidth}
\begin{tabular}{l l c c c}
\toprule
\textbf{Benchmark} & \textbf{Variant}
& \textbf{\textsc{BestScore@}\boldmath$0.5B$}
& \textbf{\textsc{BestScore@}\boldmath$B$}
& \textbf{Norm. \textsc{AUC} $\uparrow$} \\
\midrule
\multirow{4}{*}{Circle Packing $\uparrow$}
& CostAda & \textbf{\meanstd{2.6195 $\pm$ 0.0128}} & \textbf{\meanstd{2.6281 $\pm$ 0.0067}} & \textbf{\meanstd{0.9517 $\pm$ 0.0095}} \\
& w/o cost calibration & \meanstd{2.5389 $\pm$ 0.1153} & \meanstd{2.5738 $\pm$ 0.0554} & \meanstd{0.9053 $\pm$ 0.0376} \\
& w/o budget conditioning & \meanstd{2.6074 $\pm$ 0.0125} & \meanstd{2.6172 $\pm$ 0.0057} & \meanstd{0.9161 $\pm$ 0.0316} \\
& w/o intervention gating & \meanstd{2.3594 $\pm$ 0.2251} & \meanstd{2.5291 $\pm$ 0.0949} & \meanstd{0.8478 $\pm$ 0.0642} \\
\midrule
\multirow{4}{*}{Heilbronn Triangle $\uparrow$}
& CostAda & \textbf{\meanstd{0.0306 $\pm$ 0.0009}} & \textbf{\meanstd{0.0313 $\pm$ 0.0004}} & \textbf{\meanstd{0.7831 $\pm$ 0.0486}} \\
& w/o cost calibration & \meanstd{0.0167 $\pm$ 0.0032} & \meanstd{0.0228 $\pm$ 0.0040} & \meanstd{0.4654 $\pm$ 0.0614} \\
& w/o budget conditioning & \meanstd{0.0212 $\pm$ 0.0048} & \meanstd{0.0251 $\pm$ 0.0049} & \meanstd{0.4073 $\pm$ 0.0426} \\
& w/o intervention gating & \meanstd{0.0206 $\pm$ 0.0085} & \meanstd{0.0244 $\pm$ 0.0035} & \meanstd{0.4494 $\pm$ 0.1622} \\
\midrule
\multirow{4}{*}{Signal Processing $\uparrow$}
& CostAda & \textbf{\meanstd{0.6924 $\pm$ 0.0536}} & \textbf{\meanstd{0.7222 $\pm$ 0.0127}} & \textbf{\meanstd{0.6546 $\pm$ 0.0444}} \\
& w/o cost calibration & \meanstd{0.6588 $\pm$ 0.0254} & \meanstd{0.6851 $\pm$ 0.0331} & \meanstd{0.6355 $\pm$ 0.0200} \\
& w/o budget conditioning & \meanstd{0.6322 $\pm$ 0.0130} & \meanstd{0.6470 $\pm$ 0.0029} & \meanstd{0.6146 $\pm$ 0.0134} \\
& w/o intervention gating & \meanstd{0.6763 $\pm$ 0.0059} & \meanstd{0.7059 $\pm$ 0.0220} & \meanstd{0.6293 $\pm$ 0.0140} \\
\bottomrule
\end{tabular}
\end{adjustbox}
\end{center}
\end{table}

\begin{table}[!htbp]
\caption{
\textbf{Ablation study of CostAda under GPT-5.4.}
Entries report \textbf{Mean $\pm$ Std}. Benchmark objectives are used for the two cutoff columns, and higher normalized \textsc{AUC} is better. Arrows indicate the preferred direction. Best entries per benchmark and metric among CostAda and its variants are bolded, including ties at the reported precision.
}
\label{tab:gpt54-component-ablations}
\begin{center}
\begin{adjustbox}{max width=\linewidth}
\begin{tabular}{l l c c c}
\toprule
\textbf{Benchmark} & \textbf{Variant}
& \textbf{\textsc{BestScore@}\boldmath$0.5B$}
& \textbf{\textsc{BestScore@}\boldmath$B$}
& \textbf{Norm. \textsc{AUC} $\uparrow$} \\
\midrule
\multirow{4}{*}{Circle Packing $\uparrow$}
& CostAda & \textbf{\meanstd{2.6197 $\pm$ 0.0080}} & \textbf{\meanstd{2.6212 $\pm$ 0.0073}} & \textbf{\meanstd{0.9825 $\pm$ 0.0050}} \\
& w/o cost calibration & \meanstd{2.5963 $\pm$ 0.0293} & \meanstd{2.6047 $\pm$ 0.0364} & \meanstd{0.9648 $\pm$ 0.0276} \\
& w/o budget conditioning & \meanstd{2.5898 $\pm$ 0.0421} & \meanstd{2.5942 $\pm$ 0.0458} & \meanstd{0.9673 $\pm$ 0.0079} \\
& w/o intervention gating & \meanstd{2.5183 $\pm$ 0.0455} & \meanstd{2.5223 $\pm$ 0.0386} & \meanstd{0.9402 $\pm$ 0.0246} \\
\midrule
\multirow{4}{*}{Heilbronn Triangle $\uparrow$}
& CostAda & \textbf{\meanstd{0.0309 $\pm$ 0.0010}} & \textbf{\meanstd{0.0326 $\pm$ 0.0008}} & \textbf{\meanstd{0.7951 $\pm$ 0.0253}} \\
& w/o cost calibration & \meanstd{0.0252 $\pm$ 0.0021} & \meanstd{0.0274 $\pm$ 0.0018} & \meanstd{0.6748 $\pm$ 0.0459} \\
& w/o budget conditioning & \meanstd{0.0290 $\pm$ 0.0033} & \meanstd{0.0291 $\pm$ 0.0033} & \meanstd{0.7494 $\pm$ 0.0615} \\
& w/o intervention gating & \meanstd{0.0266 $\pm$ 0.0009} & \meanstd{0.0268 $\pm$ 0.0007} & \meanstd{0.6949 $\pm$ 0.0259} \\
\midrule
\multirow{4}{*}{Signal Processing $\uparrow$}
& CostAda & \textbf{\meanstd{0.7289 $\pm$ 0.0262}} & \textbf{\meanstd{0.7672 $\pm$ 0.0292}} & \textbf{\meanstd{0.7022 $\pm$ 0.0224}} \\
& w/o cost calibration & \meanstd{0.6617 $\pm$ 0.0085} & \meanstd{0.7059 $\pm$ 0.0191} & \meanstd{0.6333 $\pm$ 0.0028} \\
& w/o budget conditioning & \meanstd{0.6889 $\pm$ 0.0240} & \meanstd{0.7086 $\pm$ 0.0321} & \meanstd{0.6557 $\pm$ 0.0192} \\
& w/o intervention gating & \meanstd{0.6931 $\pm$ 0.0306} & \meanstd{0.7139 $\pm$ 0.0156} & \meanstd{0.6717 $\pm$ 0.0116} \\
\bottomrule
\end{tabular}
\end{adjustbox}
\end{center}
\end{table}

Tables~\ref{tab:glm5-component-ablations} and~\ref{tab:gpt54-component-ablations} show that the \textsc{AUC} gaps in Table~\ref{tab:component-ablation-auc} extend to the benchmark objectives at both cutoffs. Every displayed ablation result is lower than CostAda under both backbones. In particular, the lower cutoff objectives and \textsc{AUC} of w/o intervention gating occur while guidance remains available, tying this comparison to when guides are purchased and tested. On Signal Processing under GLM-5, the w/o cost calibration values are $4.86\%$, $5.13\%$, and $2.91\%$ below CostAda for \textsc{BestScore@}$0.5B$, \textsc{BestScore@}$B$, and normalized \textsc{AUC}, respectively. Under GLM-5, removing remaining-budget conditioning lowers the final Signal Processing objective below the strongest baseline mean in Table~\ref{tab:costada-math8-final-quality}. Under \mbox{GPT-5.4}, removing either cost calibration or remaining-budget conditioning also lowers it below the strongest baseline mean.

\section{Normalized Score Results}
\label{app:normalized_results}

Benchmark objectives differ in units, magnitude, and direction across the eight benchmarks. We therefore use a normalized score for the cross-benchmark cutoff and \textsc{AUC} analyses in Section~\ref{subsec:budget_cutoffs}. The shared evaluator reports this direction-adjusted quality measure for every candidate. For benchmarks with a fixed reference value, the normalized score is the ratio between the achieved objective and that reference. We invert the ratio for minimization benchmarks so that higher is always better. We adopt the reference values reported in prior work \citep{novikov2025alphaevolve,cemri2026adaevolve,liu2026evox}. A value of $1$ matches the reference, values above $1$ exceed it, and invalid candidates receive a score of zero. For Signal Processing, the evaluator already produces a bounded composite quality measure, so we use its overall program score in $[0,1]$. Normalization constants are fixed per benchmark and shared by all methods. For individual candidates, normalization preserves each benchmark's native preference ordering.

Table~\ref{tab:normalized-auc} reports per-benchmark normalized \textsc{AUC} over the full search budget, which underlies the \textsc{AUC} comparison in Section~\ref{subsec:budget_cutoffs}. Tables~\ref{tab:glm5-normalized-cutoffs} and~\ref{tab:gpt54-normalized-cutoffs} report best-so-far normalized scores at the same cutoffs as the benchmark-objective tables.

\begin{table}[!htbp]
\caption{\textbf{Normalized \textsc{AUC} over the full search budget.} Entries report \textbf{Mean $\pm$ Std} of the average best-so-far normalized score over $[0,B]$, with $B{=}1$ for GLM-5 and $B{=}5$ for GPT-5.4. Higher is better. Best entries per benchmark and backbone are bolded, including ties at the reported precision. The Average row averages the eight per-benchmark entries.}
\label{tab:normalized-auc}
\begin{center}
\begin{adjustbox}{max width=\linewidth}
\begin{tabular}{l ccc ccc}
\toprule
\multirow{2}{*}{\textbf{Benchmark}}
& \multicolumn{3}{c}{\textbf{GLM-5 (\boldmath$B{=}1$)}}
& \multicolumn{3}{c}{\textbf{GPT-5.4 (\boldmath$B{=}5$)}} \\
\cmidrule(lr){2-4}\cmidrule(lr){5-7}
& \textbf{AdaEvolve} & \textbf{EvoX} & \textbf{CostAda}
& \textbf{AdaEvolve} & \textbf{EvoX} & \textbf{CostAda} \\
\midrule
Circle Packing & \meanstd{0.8341 $\pm$ 0.0273} & \meanstd{0.8752 $\pm$ 0.0289} & \textbf{\meanstd{0.9295 $\pm$ 0.0134}} & \meanstd{0.9668 $\pm$ 0.0197} & \meanstd{0.9548 $\pm$ 0.0172} & \textbf{\meanstd{0.9771 $\pm$ 0.0096}} \\
Circle Packing Rect & \meanstd{0.9009 $\pm$ 0.0629} & \meanstd{0.8386 $\pm$ 0.1738} & \textbf{\meanstd{0.9551 $\pm$ 0.0133}} & \meanstd{0.9630 $\pm$ 0.0139} & \meanstd{0.9668 $\pm$ 0.0124} & \textbf{\meanstd{0.9672 $\pm$ 0.0058}} \\
Heilbronn Convex & \meanstd{0.6812 $\pm$ 0.0840} & \meanstd{0.6579 $\pm$ 0.0560} & \textbf{\meanstd{0.7606 $\pm$ 0.0700}} & \meanstd{0.5896 $\pm$ 0.0243} & \meanstd{0.6103 $\pm$ 0.0645} & \textbf{\meanstd{0.7499 $\pm$ 0.0144}} \\
Heilbronn Triangle & \meanstd{0.6571 $\pm$ 0.0658} & \meanstd{0.4700 $\pm$ 0.0372} & \textbf{\meanstd{0.7613 $\pm$ 0.0531}} & \meanstd{0.7825 $\pm$ 0.0132} & \meanstd{0.6463 $\pm$ 0.0632} & \textbf{\meanstd{0.8229 $\pm$ 0.0387}} \\
MinMaxDist $(n{=}16,d{=}2)$ & \meanstd{0.8864 $\pm$ 0.0588} & \meanstd{0.8530 $\pm$ 0.0907} & \textbf{\meanstd{0.9241 $\pm$ 0.0370}} & \meanstd{0.9556 $\pm$ 0.0343} & \meanstd{0.9651 $\pm$ 0.0165} & \textbf{\meanstd{0.9654 $\pm$ 0.0098}} \\
MinMaxDist $(n{=}14,d{=}3)$ & \meanstd{0.9172 $\pm$ 0.0068} & \meanstd{0.9232 $\pm$ 0.0473} & \textbf{\meanstd{0.9502 $\pm$ 0.0082}} & \meanstd{0.8947 $\pm$ 0.0240} & \meanstd{0.9031 $\pm$ 0.0041} & \textbf{\meanstd{0.9186 $\pm$ 0.0077}} \\
Third Autocorrelation & \textbf{\meanstd{0.9820 $\pm$ 0.0027}} & \meanstd{0.9177 $\pm$ 0.1009} & \meanstd{0.9658 $\pm$ 0.0050} & \meanstd{0.9311 $\pm$ 0.0102} & \meanstd{0.9609 $\pm$ 0.0041} & \textbf{\meanstd{0.9776 $\pm$ 0.0081}} \\
Signal Processing & \meanstd{0.5891 $\pm$ 0.0149} & \meanstd{0.5873 $\pm$ 0.0366} & \textbf{\meanstd{0.6546 $\pm$ 0.0444}} & \meanstd{0.6660 $\pm$ 0.0024} & \meanstd{0.6216 $\pm$ 0.0292} & \textbf{\meanstd{0.7022 $\pm$ 0.0224}} \\
\midrule
Average & 0.8060 & 0.7654 & \textbf{0.8626} & 0.8437 & 0.8286 & \textbf{0.8851} \\
\bottomrule
\end{tabular}
\end{adjustbox}
\end{center}
\end{table}

CostAda achieves the highest normalized \textsc{AUC} on fifteen of the sixteen benchmark--backbone pairs in Table~\ref{tab:normalized-auc}. Its eight-benchmark average is also the highest under GLM-5 and \mbox{GPT-5.4}, at $0.8626$ and $0.8851$ compared with $0.8060$ and $0.8437$ for AdaEvolve. CostAda's cutoff-cell rankings also agree between normalized-score and benchmark-objective reporting. In Tables~\ref{tab:glm5-normalized-cutoffs} and~\ref{tab:gpt54-normalized-cutoffs}, CostAda is best or tied in 30 of the 32 cells under GLM-5 and in 29 under \mbox{GPT-5.4}, the same counts as in Appendix~\ref{app:additional_budget_cutoffs}. This agreement spans both maximization and minimization benchmarks after their score directions are aligned, so normalized scores support the aggregate comparison.

\begin{table}[!htbp]
\caption{\textbf{Best-so-far normalized scores across GLM-5 budget cutoffs.} The search budget is $B{=}1$, with the same budget fractions as Table~\ref{tab:glm5-budget-cutoffs}. Entries report \textbf{Mean $\pm$ Std}. Higher is better. Best entries per benchmark and cutoff are bolded, including ties at the reported precision.}
\label{tab:glm5-normalized-cutoffs}
\begin{center}
\begin{adjustbox}{max width=\linewidth}
\begin{tabular}{l l c c c c}
\toprule
\textbf{Benchmark} & \textbf{Method} & \textbf{\boldmath$0.25B$} & \textbf{\boldmath$0.5B$} & \textbf{\boldmath$0.75B$} & \textbf{\boldmath$B$} \\
\midrule
\multirow{3}{*}{Circle Packing}
& AdaEvolve & \meanstd{0.6917 $\pm$ 0.0844} & \meanstd{0.9370 $\pm$ 0.0593} & \meanstd{0.9722 $\pm$ 0.0293} & \meanstd{0.9722 $\pm$ 0.0293} \\
& EvoX & \meanstd{0.8279 $\pm$ 0.0304} & \meanstd{0.9536 $\pm$ 0.0481} & \meanstd{0.9760 $\pm$ 0.0285} & \meanstd{0.9760 $\pm$ 0.0285} \\
& CostAda & \textbf{\meanstd{0.9691 $\pm$ 0.0222}} & \textbf{\meanstd{0.9873 $\pm$ 0.0127}} & \textbf{\meanstd{0.9963 $\pm$ 0.0036}} & \textbf{\meanstd{0.9983 $\pm$ 0.0021}} \\
\midrule
\multirow{3}{*}{Circle Packing Rect}
& AdaEvolve & \meanstd{0.8953 $\pm$ 0.0484} & \meanstd{0.9317 $\pm$ 0.0667} & \meanstd{0.9336 $\pm$ 0.0691} & \meanstd{0.9336 $\pm$ 0.0691} \\
& EvoX & \meanstd{0.9293 $\pm$ 0.1086} & \meanstd{0.9295 $\pm$ 0.1083} & \meanstd{0.9295 $\pm$ 0.1083} & \meanstd{0.9295 $\pm$ 0.1083} \\
& CostAda & \textbf{\meanstd{0.9924 $\pm$ 0.0034}} & \textbf{\meanstd{0.9924 $\pm$ 0.0034}} & \textbf{\meanstd{0.9924 $\pm$ 0.0034}} & \textbf{\meanstd{0.9944 $\pm$ 0.0034}} \\
\midrule
\multirow{3}{*}{Heilbronn Convex}
& AdaEvolve & \meanstd{0.6954 $\pm$ 0.0963} & \meanstd{0.6954 $\pm$ 0.0963} & \meanstd{0.7410 $\pm$ 0.0497} & \meanstd{0.7410 $\pm$ 0.0497} \\
& EvoX & \meanstd{0.6556 $\pm$ 0.1173} & \meanstd{0.7166 $\pm$ 0.0790} & \meanstd{0.7563 $\pm$ 0.0372} & \meanstd{0.7608 $\pm$ 0.0386} \\
& CostAda & \textbf{\meanstd{0.6957 $\pm$ 0.0104}} & \textbf{\meanstd{0.8096 $\pm$ 0.0974}} & \textbf{\meanstd{0.8190 $\pm$ 0.1078}} & \textbf{\meanstd{0.8496 $\pm$ 0.0575}} \\
\midrule
\multirow{3}{*}{Heilbronn Triangle}
& AdaEvolve & \meanstd{0.5816 $\pm$ 0.2042} & \meanstd{0.7734 $\pm$ 0.0535} & \meanstd{0.7734 $\pm$ 0.0535} & \meanstd{0.8300 $\pm$ 0.0309} \\
& EvoX & \meanstd{0.4088 $\pm$ 0.0730} & \meanstd{0.5488 $\pm$ 0.1039} & \meanstd{0.5488 $\pm$ 0.1039} & \meanstd{0.6024 $\pm$ 0.1364} \\
& CostAda & \textbf{\meanstd{0.8098 $\pm$ 0.0452}} & \textbf{\meanstd{0.8098 $\pm$ 0.0452}} & \textbf{\meanstd{0.8098 $\pm$ 0.0452}} & \textbf{\meanstd{0.8595 $\pm$ 0.0085}} \\
\midrule
\multirow{3}{*}{MinMaxDist $(n{=}16,d{=}2)$}
& AdaEvolve & \meanstd{0.8861 $\pm$ 0.0630} & \meanstd{0.8891 $\pm$ 0.0626} & \meanstd{0.9214 $\pm$ 0.0811} & \meanstd{0.9214 $\pm$ 0.0811} \\
& EvoX & \meanstd{0.8657 $\pm$ 0.0957} & \meanstd{0.9090 $\pm$ 0.1232} & \meanstd{0.9179 $\pm$ 0.1138} & \meanstd{0.9201 $\pm$ 0.1156} \\
& CostAda & \textbf{\meanstd{0.9495 $\pm$ 0.0681}} & \textbf{\meanstd{0.9540 $\pm$ 0.0721}} & \textbf{\meanstd{0.9805 $\pm$ 0.0264}} & \textbf{\meanstd{0.9838 $\pm$ 0.0212}} \\
\midrule
\multirow{3}{*}{MinMaxDist $(n{=}14,d{=}3)$}
& AdaEvolve & \meanstd{0.9119 $\pm$ 0.0170} & \meanstd{0.9298 $\pm$ 0.0264} & \meanstd{0.9512 $\pm$ 0.0168} & \meanstd{0.9512 $\pm$ 0.0168} \\
& EvoX & \meanstd{0.9120 $\pm$ 0.0766} & \meanstd{0.9612 $\pm$ 0.0568} & \meanstd{0.9864 $\pm$ 0.0142} & \meanstd{0.9909 $\pm$ 0.0076} \\
& CostAda & \textbf{\meanstd{0.9685 $\pm$ 0.0522}} & \textbf{\meanstd{0.9936 $\pm$ 0.0094}} & \textbf{\meanstd{0.9964 $\pm$ 0.0045}} & \textbf{\meanstd{0.9994 $\pm$ 0.0006}} \\
\midrule
\multirow{3}{*}{Third Autocorrelation}
& AdaEvolve & \textbf{\meanstd{0.9895 $\pm$ 0.0027}} & \textbf{\meanstd{0.9907 $\pm$ 0.0011}} & \meanstd{0.9907 $\pm$ 0.0011} & \meanstd{0.9913 $\pm$ 0.0009} \\
& EvoX & \meanstd{0.9307 $\pm$ 0.0960} & \meanstd{0.9326 $\pm$ 0.0927} & \meanstd{0.9446 $\pm$ 0.0761} & \meanstd{0.9446 $\pm$ 0.0761} \\
& CostAda & \meanstd{0.9799 $\pm$ 0.0167} & \meanstd{0.9871 $\pm$ 0.0043} & \textbf{\meanstd{0.9916 $\pm$ 0.0036}} & \textbf{\meanstd{0.9931 $\pm$ 0.0025}} \\
\midrule
\multirow{3}{*}{Signal Processing}
& AdaEvolve & \meanstd{0.5692 $\pm$ 0.0301} & \meanstd{0.6126 $\pm$ 0.0161} & \meanstd{0.6173 $\pm$ 0.0158} & \meanstd{0.6393 $\pm$ 0.0240} \\
& EvoX & \meanstd{0.5122 $\pm$ 0.0177} & \meanstd{0.6361 $\pm$ 0.0655} & \meanstd{0.6429 $\pm$ 0.0544} & \meanstd{0.6502 $\pm$ 0.0430} \\
& CostAda & \textbf{\meanstd{0.6377 $\pm$ 0.0805}} & \textbf{\meanstd{0.6924 $\pm$ 0.0536}} & \textbf{\meanstd{0.7069 $\pm$ 0.0318}} & \textbf{\meanstd{0.7222 $\pm$ 0.0127}} \\
\bottomrule
\end{tabular}
\end{adjustbox}
\end{center}
\end{table}

\clearpage
\begin{table}[H]
\caption{\textbf{Best-so-far normalized scores across GPT-5.4 budget cutoffs.} The search budget is $B{=}5$, with the same budget fractions as Table~\ref{tab:gpt54-budget-cutoffs}. Entries report \textbf{Mean $\pm$ Std}. Higher is better. Best entries per benchmark and cutoff are bolded, including ties at the reported precision.}
\label{tab:gpt54-normalized-cutoffs}
\begin{center}
\begin{adjustbox}{max width=\linewidth}
\begin{tabular}{l l c c c c}
\toprule
\textbf{Benchmark} & \textbf{Method} & \textbf{\boldmath$0.25B$} & \textbf{\boldmath$0.5B$} & \textbf{\boldmath$0.75B$} & \textbf{\boldmath$B$} \\
\midrule
\multirow{3}{*}{Circle Packing}
& AdaEvolve & \meanstd{0.9727 $\pm$ 0.0297} & \meanstd{0.9801 $\pm$ 0.0194} & \meanstd{0.9816 $\pm$ 0.0203} & \meanstd{0.9941 $\pm$ 0.0011} \\
& EvoX & \meanstd{0.9573 $\pm$ 0.0125} & \meanstd{0.9655 $\pm$ 0.0231} & \meanstd{0.9798 $\pm$ 0.0181} & \meanstd{0.9811 $\pm$ 0.0193} \\
& CostAda & \textbf{\meanstd{0.9921 $\pm$ 0.0053}} & \textbf{\meanstd{0.9947 $\pm$ 0.0012}} & \textbf{\meanstd{0.9950 $\pm$ 0.0017}} & \textbf{\meanstd{0.9961 $\pm$ 0.0017}} \\
\midrule
\multirow{3}{*}{Circle Packing Rect}
& AdaEvolve & \meanstd{0.9648 $\pm$ 0.0173} & \meanstd{0.9684 $\pm$ 0.0214} & \meanstd{0.9775 $\pm$ 0.0156} & \meanstd{0.9784 $\pm$ 0.0151} \\
& EvoX & \textbf{\meanstd{0.9815 $\pm$ 0.0181}} & \textbf{\meanstd{0.9922 $\pm$ 0.0004}} & \meanstd{0.9922 $\pm$ 0.0004} & \meanstd{0.9922 $\pm$ 0.0004} \\
& CostAda & \meanstd{0.9590 $\pm$ 0.0132} & \meanstd{0.9769 $\pm$ 0.0195} & \textbf{\meanstd{0.9929 $\pm$ 0.0054}} & \textbf{\meanstd{0.9966 $\pm$ 0.0021}} \\
\midrule
\multirow{3}{*}{Heilbronn Convex}
& AdaEvolve & \meanstd{0.6034 $\pm$ 0.0242} & \meanstd{0.6034 $\pm$ 0.0242} & \meanstd{0.6049 $\pm$ 0.0222} & \meanstd{0.6376 $\pm$ 0.0640} \\
& EvoX & \meanstd{0.5353 $\pm$ 0.0594} & \meanstd{0.6354 $\pm$ 0.0829} & \meanstd{0.6840 $\pm$ 0.0663} & \meanstd{0.7194 $\pm$ 0.1014} \\
& CostAda & \textbf{\meanstd{0.7574 $\pm$ 0.0319}} & \textbf{\meanstd{0.7574 $\pm$ 0.0319}} & \textbf{\meanstd{0.7885 $\pm$ 0.0243}} & \textbf{\meanstd{0.7885 $\pm$ 0.0243}} \\
\midrule
\multirow{3}{*}{Heilbronn Triangle}
& AdaEvolve & \meanstd{0.7654 $\pm$ 0.0088} & \meanstd{0.7941 $\pm$ 0.0425} & \meanstd{0.8183 $\pm$ 0.0382} & \meanstd{0.8395 $\pm$ 0.0219} \\
& EvoX & \meanstd{0.6577 $\pm$ 0.0599} & \meanstd{0.6940 $\pm$ 0.0507} & \meanstd{0.7124 $\pm$ 0.0824} & \meanstd{0.7205 $\pm$ 0.0956} \\
& CostAda & \textbf{\meanstd{0.7897 $\pm$ 0.1072}} & \textbf{\meanstd{0.8675 $\pm$ 0.0376}} & \textbf{\meanstd{0.8997 $\pm$ 0.0334}} & \textbf{\meanstd{0.9042 $\pm$ 0.0258}} \\
\midrule
\multirow{3}{*}{MinMaxDist $(n{=}16,d{=}2)$}
& AdaEvolve & \meanstd{0.9579 $\pm$ 0.0582} & \meanstd{0.9653 $\pm$ 0.0453} & \meanstd{0.9811 $\pm$ 0.0180} & \meanstd{0.9914 $\pm$ 0.0002} \\
& EvoX & \textbf{\meanstd{0.9915 $\pm$ 0.0000}} & \meanstd{0.9915 $\pm$ 0.0000} & \meanstd{0.9915 $\pm$ 0.0000} & \meanstd{0.9915 $\pm$ 0.0000} \\
& CostAda & \meanstd{0.9796 $\pm$ 0.0149} & \textbf{\meanstd{0.9929 $\pm$ 0.0065}} & \textbf{\meanstd{0.9972 $\pm$ 0.0049}} & \textbf{\meanstd{0.9972 $\pm$ 0.0049}} \\
\midrule
\multirow{3}{*}{MinMaxDist $(n{=}14,d{=}3)$}
& AdaEvolve & \meanstd{0.8980 $\pm$ 0.0305} & \meanstd{0.9045 $\pm$ 0.0267} & \meanstd{0.9045 $\pm$ 0.0267} & \meanstd{0.9156 $\pm$ 0.0305} \\
& EvoX & \meanstd{0.8991 $\pm$ 0.0325} & \meanstd{0.9347 $\pm$ 0.0018} & \meanstd{0.9376 $\pm$ 0.0050} & \meanstd{0.9376 $\pm$ 0.0050} \\
& CostAda & \textbf{\meanstd{0.9156 $\pm$ 0.0305}} & \textbf{\meanstd{0.9434 $\pm$ 0.0177}} & \textbf{\meanstd{0.9434 $\pm$ 0.0177}} & \textbf{\meanstd{0.9434 $\pm$ 0.0177}} \\
\midrule
\multirow{3}{*}{Third Autocorrelation}
& AdaEvolve & \meanstd{0.9434 $\pm$ 0.0169} & \meanstd{0.9604 $\pm$ 0.0050} & \meanstd{0.9604 $\pm$ 0.0050} & \meanstd{0.9652 $\pm$ 0.0044} \\
& EvoX & \meanstd{0.9770 $\pm$ 0.0018} & \meanstd{0.9801 $\pm$ 0.0028} & \meanstd{0.9810 $\pm$ 0.0019} & \meanstd{0.9820 $\pm$ 0.0028} \\
& CostAda & \textbf{\meanstd{0.9870 $\pm$ 0.0038}} & \textbf{\meanstd{0.9877 $\pm$ 0.0032}} & \textbf{\meanstd{0.9891 $\pm$ 0.0050}} & \textbf{\meanstd{0.9891 $\pm$ 0.0050}} \\
\midrule
\multirow{3}{*}{Signal Processing}
& AdaEvolve & \meanstd{0.6617 $\pm$ 0.0040} & \meanstd{0.6739 $\pm$ 0.0145} & \meanstd{0.7043 $\pm$ 0.0024} & \meanstd{0.7182 $\pm$ 0.0137} \\
& EvoX & \meanstd{0.5890 $\pm$ 0.0447} & \meanstd{0.6436 $\pm$ 0.0509} & \meanstd{0.6644 $\pm$ 0.0299} & \meanstd{0.6767 $\pm$ 0.0179} \\
& CostAda & \textbf{\meanstd{0.7069 $\pm$ 0.0402}} & \textbf{\meanstd{0.7289 $\pm$ 0.0262}} & \textbf{\meanstd{0.7289 $\pm$ 0.0262}} & \textbf{\meanstd{0.7672 $\pm$ 0.0292}} \\
\bottomrule
\end{tabular}
\end{adjustbox}
\end{center}
\end{table}

\FloatBarrier

\section{Budget Adherence}
\label{app:budget_adherence}

Final realized spending remains close to the nominal budgets across methods and backbones (\mbox{Table~\ref{tab:budget-adherence}}). We report final spending as \textsc{AvgCost} and the relative excess as \textsc{OvershootRatio}. The average excess ranges from $1.1\%$ to $2.5\%$, and no run exceeds the budget by more than $8.6\%$. Moreover, CostAda has the smallest average final spending and overshoot under GLM-5. The full-budget quality comparison in Section~\ref{subsec:final_quality} therefore uses closely matched realized spending.

\begin{table}[!htbp]
\caption{\textbf{Final realized spending and budget overshoot.} \textsc{AvgCost} reports \textbf{Mean $\pm$ Std} of final cumulative search-side cost. \textsc{OvershootRatio} is the relative excess over the nominal budget. For each backbone, statistics aggregate all eight benchmarks (24 runs per method). Entries are not ranked, since spending alone does not measure solution quality.}
\label{tab:budget-adherence}
\begin{center}
\begin{adjustbox}{max width=\linewidth}
\begin{tabular}{l l c c c}
\toprule
\textbf{Backbone} & \textbf{Method} & \textbf{\textsc{AvgCost}} & \textbf{Mean \textsc{OvershootRatio}} & \textbf{Max \textsc{OvershootRatio}} \\
\midrule
\multirow{3}{*}{GLM-5 ($B{=}1$)}
& AdaEvolve & \meanstd[0.000]{1.021 $\pm$ 0.019} & 2.1\% & 8.0\% \\
& EvoX & \meanstd[0.000]{1.025 $\pm$ 0.022} & 2.5\% & 8.6\% \\
& CostAda & \meanstd[0.000]{1.017 $\pm$ 0.022} & 1.7\% & 7.1\% \\
\midrule
\multirow{3}{*}{GPT-5.4 ($B{=}5$)}
& AdaEvolve & \meanstd[0.000]{5.092 $\pm$ 0.068} & 1.8\% & 5.6\% \\
& EvoX & \meanstd[0.000]{5.053 $\pm$ 0.064} & 1.1\% & 6.2\% \\
& CostAda & \meanstd[0.000]{5.103 $\pm$ 0.081} & 2.1\% & 6.8\% \\
\bottomrule
\end{tabular}
\end{adjustbox}
\end{center}
\end{table}

\section{Discovered Programs}
\label{app:discovered_programs}

\begin{figure}[!htbp]
\centering
\includegraphics[width=\linewidth]{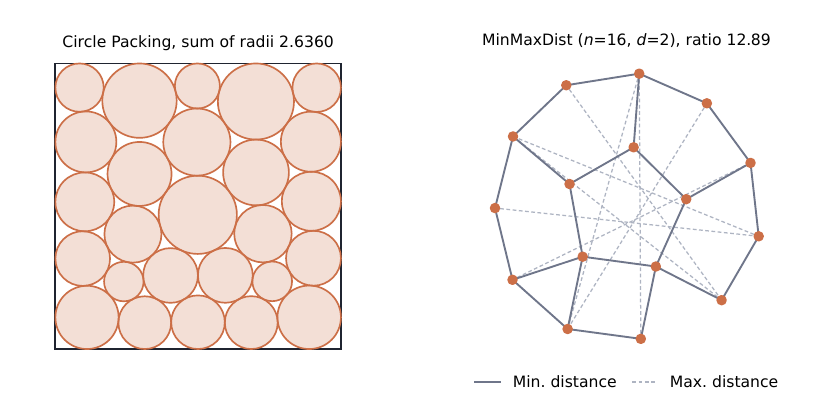}
\caption{\textbf{Configurations returned by two final CostAda programs under GLM-5.} The left panel shows the Circle Packing solution in the unit square. In the right panel, solid lines join point pairs at the minimum distance, and dashed lines join pairs at the maximum distance.}
\label{fig:discovered-configurations}
\end{figure}

Two final CostAda programs illustrate the human-reference results in Section~\ref{subsec:final_quality}. Figure~\ref{fig:discovered-configurations} shows the configurations returned by the GLM-5 programs whose final objectives are the Circle Packing and MinMaxDist $(n{=}16,d{=}2)$ Best entries in Table~\ref{tab:costada-math8-final-quality}. The Circle Packing program places 26 disjoint circles in the unit square with a sum of radii of $2.6360$, above the reported reference of $2.6340$. The MinMaxDist program returns 16 points whose maximum and minimum squared pairwise distances have a ratio of $12.89$, matching the reference value. Both are deterministic Python programs, so executing them reproduces these values.

The two programs reach these values through different uses of constrained optimization. The Circle Packing program starts from a corner-weighted layout with circles at the corners, along the edges, and in the interior. Basin hopping then perturbs this layout, and a constrained local solver (SLSQP) optimizes all centers and radii jointly. The solver's inequality constraints keep each circle inside the square and every pair of circles disjoint. The MinMaxDist program instead introduces an auxiliary variable $t$ and maximizes it subject to every squared pairwise distance lying between $t$ and one. This reformulation turns the ratio objective into a smooth constrained problem, which the program solves from hexagonal, grid, and circular initializations before keeping the best configuration. Figure~\ref{fig:discovered-program-code} lists the core of both programs, including their constraint functions and solver calls.

\begin{figure}[!htbp]
\centering
\begin{minipage}{\linewidth}
\scriptsize
\hrule\vspace{3pt}
\textbf{Circle Packing}
\begin{verbatim}
def construct_packing():
    n = 26
    best_sum, best_x = 0, None
    # [...] boundary_cons(x): circle inside the unit square; overlap_cons(x): d_ij - r_i - r_j
    for seed in [42, 123, 456]:
        np.random.seed(seed)
        centers0 = _init_config(n)
        radii0 = _compute_radii(centers0)
        x0 = np.concatenate([centers0.flatten(), radii0])

        bounds = [(0.01, 0.99)] * (2*n) + [(0.001, 0.45)] * n
        minimizer_kwargs = {
            'method': 'SLSQP',
            'bounds': bounds,
            'constraints': [{'type': 'ineq', 'fun': boundary_cons},
                          {'type': 'ineq', 'fun': overlap_cons}],
            'options': {'maxiter': 500, 'ftol': 1e-9}
        }
        result = basinhopping(
            lambda x: -np.sum(x[2*n:]), x0,
            niter=300, T=1.0, stepsize=0.08,
            minimizer_kwargs=minimizer_kwargs,
            niter_success=50
        )
        if -result.fun > best_sum:
            best_sum, best_x = -result.fun, result.x
\end{verbatim}
\hrule\vspace{3pt}
\textbf{MinMaxDist $(n{=}16,d{=}2)$}
\begin{verbatim}
def min_max_dist_dim2_16() -> np.ndarray:
    n = 16
    np.random.seed(42)
    # [...] compute_ratio(pts): min/max distance ratio of a configuration
    def objective(x):
        return -x[-1]

    def min_dist_constraints(x):
        pts = x[:-1].reshape(n, 2)
        t = x[-1]
        vals = []
        for i in range(n):
            for j in range(i + 1, n):
                d_sq = (pts[i, 0] - pts[j, 0])**2 + (pts[i, 1] - pts[j, 1])**2
                vals.append(t - d_sq)
        return np.array(vals)
    # [...] max_dist_constraints(x) returns d_sq - 1.0 for every pair
    constraints = [
        {'type': 'ineq', 'fun': lambda x: -min_dist_constraints(x)},
        {'type': 'ineq', 'fun': lambda x: -max_dist_constraints(x)}
    ]
    bounds = [(0.0, 1.0)] * (2 * n) + [(0.01, 0.5)]
    # [...] init_hexagonal, init_grid, init_circular
    best_pts, best_ratio = None, -1.0
    for init_func in [init_hexagonal, init_grid, init_circular]:
        pts0 = init_func()
        # [...] t0 is 0.9 times the smallest initial squared distance
        x0 = np.concatenate([pts0.flatten(), [t0]])
        result = minimize(objective, x0, method='SLSQP', bounds=bounds,
                         constraints=constraints,
                         options={'ftol': 1e-9, 'maxiter': 1000})
        # [...] keep the configuration with the largest distance ratio
\end{verbatim}
\hrule
\end{minipage}
\caption{\textbf{Core of the two discovered programs.} Code lines are verbatim excerpts from the final programs, shown without their comments and docstrings. Comments beginning with \texttt{[...]} mark omitted code, such as helper functions and initializations described in the text.}
\label{fig:discovered-program-code}
\end{figure}

\end{document}